\pdfoutput=1
\documentclass{article}

\usepackage[main, final]{neurips_2026}

\usepackage[utf8]{inputenc}
\usepackage[T1]{fontenc}
\usepackage{hyperref}
\usepackage{url}
\usepackage{booktabs}
\usepackage{amsfonts}
\usepackage{amsmath}
\usepackage{amssymb}
\usepackage{nicefrac}
\usepackage{microtype}
\usepackage{xcolor}
\usepackage{graphicx}
\usepackage{subcaption}
\usepackage{algorithm}
\usepackage{algorithmic}
\usepackage{multirow}
\usepackage{makecell}
\usepackage{wrapfig}
\usepackage{enumitem}
\usepackage{amsthm}

\theoremstyle{definition}
\newtheorem{definition}{Definition}

\title{Locating and Repairing Domain Shift in VLM Trajectory Planning}

\author{%
  Zhihong Cui$^{1}$ \quad Hengyu Liu$^{2}$\thanks{Corresponding author: Hengyu Liu (\texttt{heli@cs.aau.dk}).} \quad Michael Alexander Riegler$^{3}$ \\
  \textbf{Guandong Xu$^{4}$ \quad Amir Taherkordi$^{1}$ \quad Tor Skeie$^{1,5}$} \\[4pt]
  $^{1}$Department of Informatics, University of Oslo, Norway \\
  $^{2}$Department of Computer Science, Aalborg University, Denmark \\
  $^{3}$Simula Metropolitan Center for Digital Engineering (SimulaMet), Norway \\
  $^{4}$The Education University of Hong Kong, Hong Kong, China \\
  $^{5}$Simula Research Laboratory, Norway \\[2pt]
  \texttt{\{zhihongc, amirhost, tskeie\}@ifi.uio.no}, \texttt{heli@cs.aau.dk}, \\
  \texttt{michael@simula.no}, \texttt{gdxu@eduhk.hk}
}

\begin{document}

\maketitle

\begin{abstract}
Vision-Language Models (VLMs) are increasingly adopted as end-to-end driving planners, yet they suffer from cross-city domain shift. Existing methods treat this shift as a single quantity addressed by uniform alignment or single-site editing. We challenge this view on two counts: \textbf{\textit{(i) domain shift in VLM trajectory planners is inherently structured across modality and layer, and (ii) a locate-then-repair recipe matched to this structure outperforms every uniform-objective baseline}.} We analyse domain shift via activation patching and uncover two regularities. \textbf{(F1)~Modality axis.} Image and text tokens exhibit distinct shift patterns in each VLM. \textbf{(F2)~Layer axis.} The layer-wise shift partition is architecture-dependent and dataset-stable. These findings recast domain shift as a \textbf{two-dimensional tensor} indexed by \emph{modality} (image vs.\ text tokens) and \emph{layer}, motivating a simple principle: \textbf{\textit{locate where shift occurs, and repair only there}}.
We instantiate this principle in \textbf{MoLaRx} (\textbf{Mo}dality--\textbf{La}yer \textbf{R}epair), a locate-then-repair framework.
The Causal Importance Score (CIS) estimates the domain shift as a modality--layer tensor and partitions the layers into functional zones.
CIS-Guided Selective Repair (CISR) then applies a matched operator (Maximum Mean Discrepancy (MMD) / low-rank adaptation (LoRA) / freezing) to repair each zone, with the composition adapting automatically to each architecture. We evaluate \textbf{MoLaRx} on three architecturally distinct VLMs (LFM2-VL-1.6B, Qwen2-VL-2B, and PaliGemma2-3B) across two cross-city benchmarks (nuScenes and Argoverse~2). MoLaRx attains the lowest cross-city trajectory-error gap among adaptation methods in all six main-table settings, while using $50$--$60\%$ fewer trainable parameters and $8.4$--$10.9$\,ms lower inference latency. In contrast, uniform MMD-only adaptation collapses in all six settings, highlighting the importance of structure-guided repair. Code: \url{https://github.com/ZhihongCui/MoLaRx}.

\end{abstract}

\section{Introduction}
\label{sec:introduction}
\begin{figure}[htbp]
  \centering
  \includegraphics[width=\linewidth]{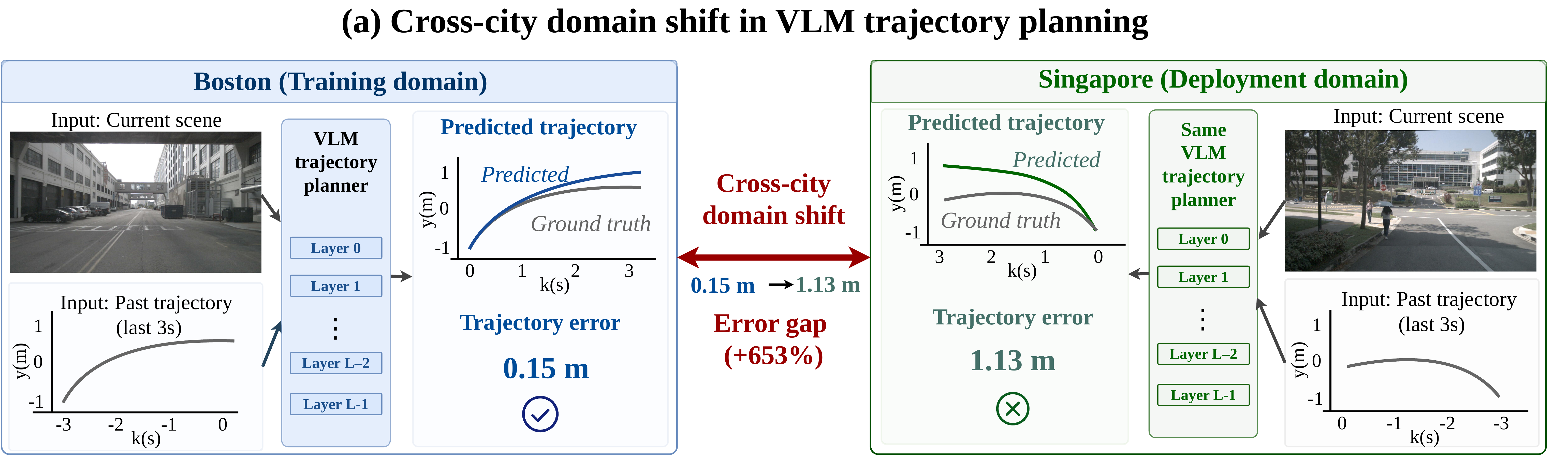}
  \includegraphics[width=0.319\linewidth]{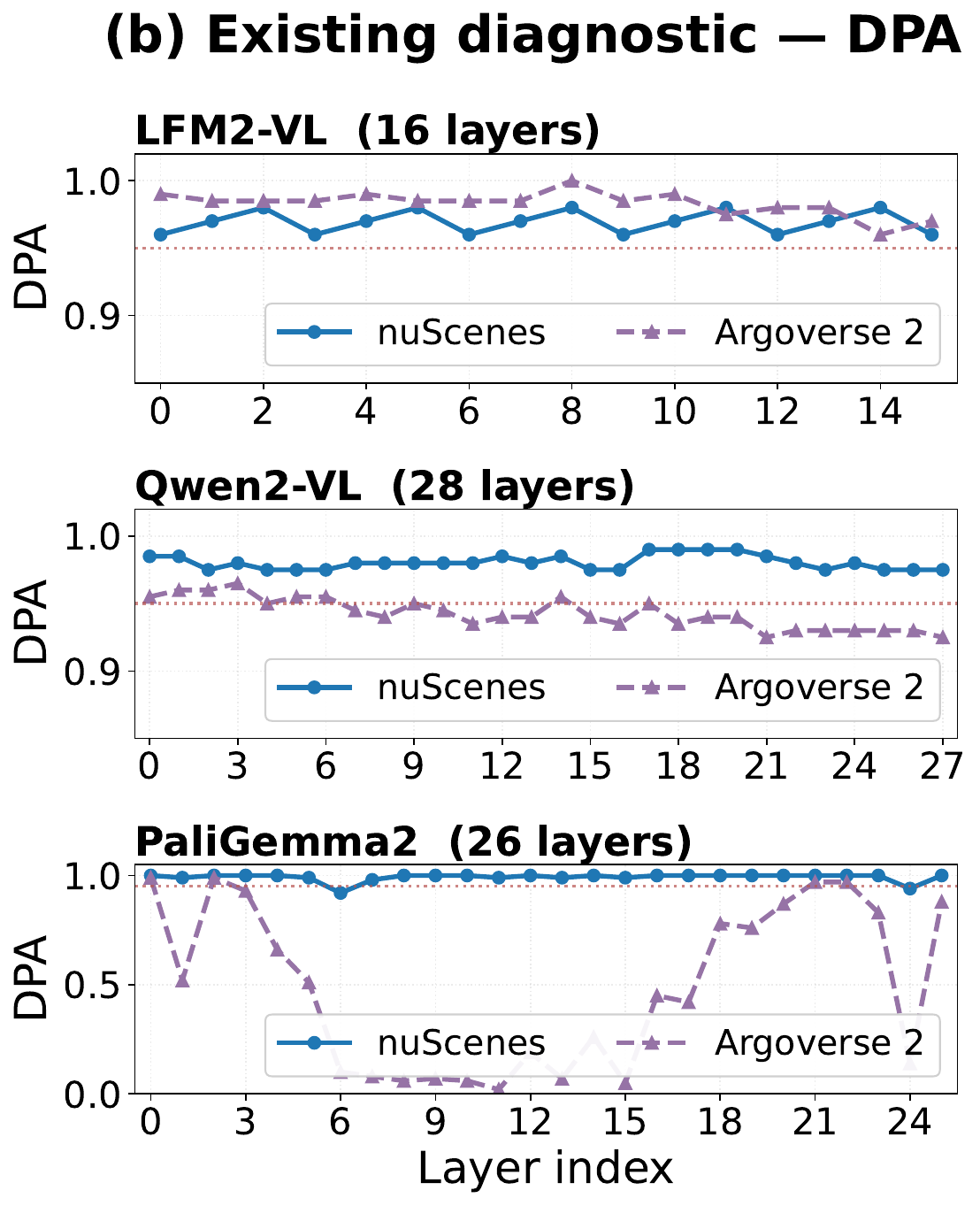}\hfill
  \includegraphics[width=0.315\linewidth]{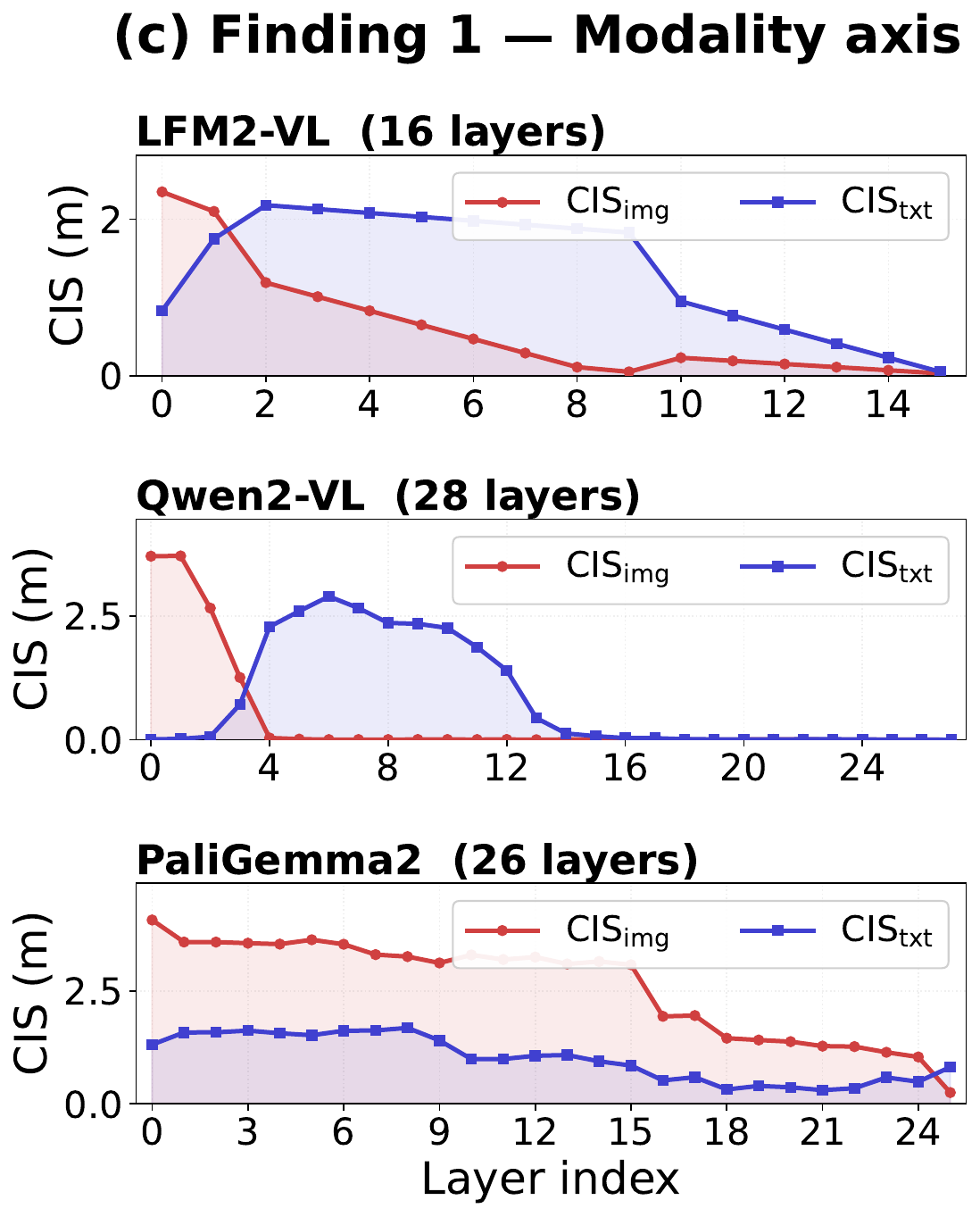}\hfill
  \includegraphics[width=0.33\linewidth]{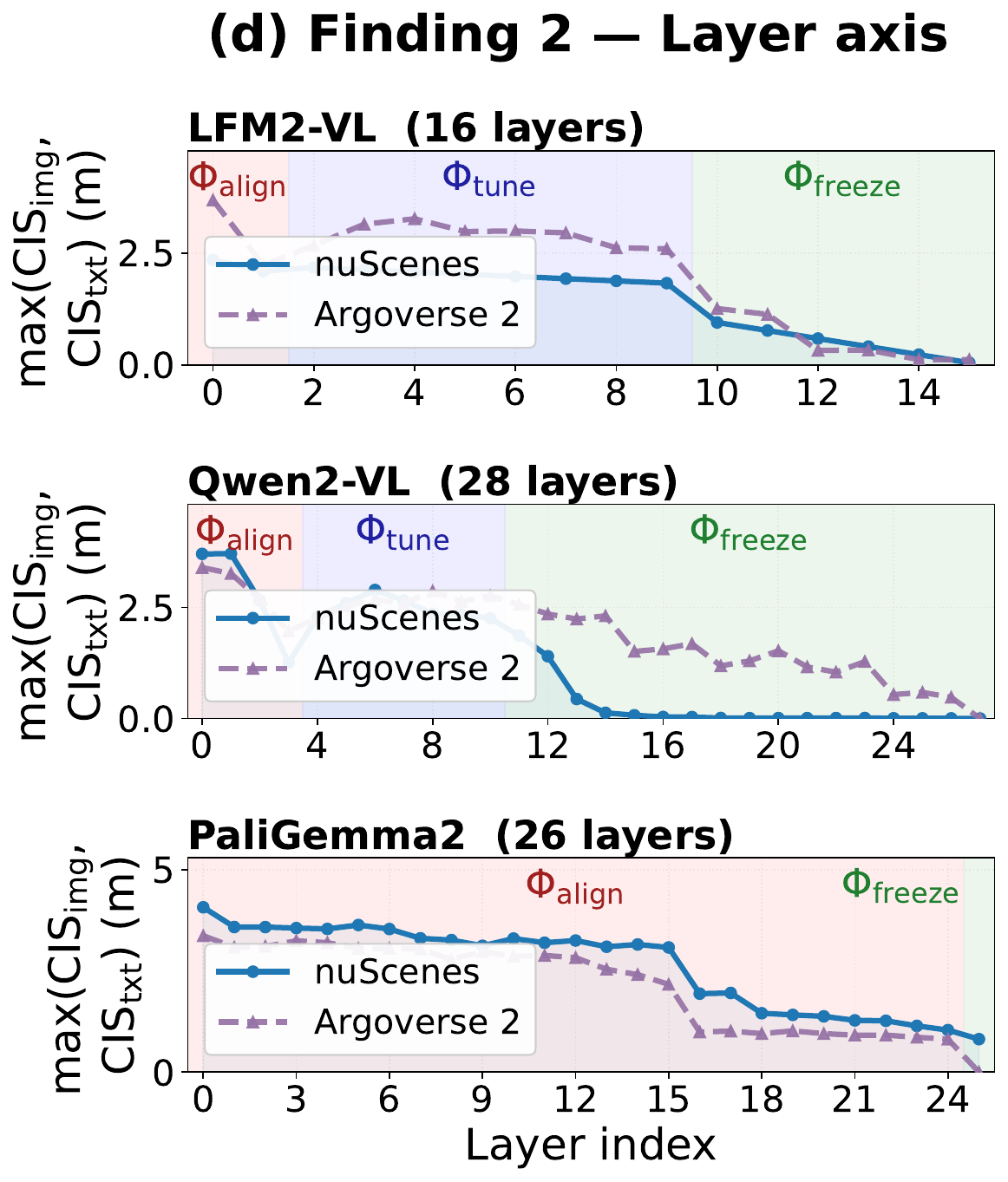}
  \caption{Domain shift in VLM trajectory planners is structured along modality and layer.
  \textbf{(a)} VLMs fine-tuned as trajectory planners suffer cross-city domain shift;
  \textbf{(b)} the existing diagnostic --- DPA: saturates near 1 at every layer of every VLM;
  \textbf{(c) F1 (modality axis)}: image and text shifts differ within every VLM;
  \textbf{(d) F2 (layer axis)}: zone partition is architecture-dependent and dataset-stable.}
  \vspace{-6mm}
  \label{fig:fig1}
\end{figure}

Vision-Language Models (VLMs) are increasingly being introduced for end-to-end trajectory planning~\citep{mao2023gptdriver, tian2024drivevlm, shao2024lmdrive, sima2024drivelm, sha2023languagempc} as they naturally integrate visual perception with historical trajectory context for decision-making.
However, they suffer from severe cross-city domain shift~\citep{naeinian2026zero}: as shown in Figure~\ref{fig:fig1} (a), a model trained in Boston incurs a $+653\%$ increase in trajectory error when deployed in Singapore (from 0.15\,m to 1.13\,m).
Since collecting and annotating data for every new city is impractical, mitigating this gap is critical for safe and scalable deployment.

Two paradigms in the broader literature shape how such cross-domain gaps are tackled in deep models.
The first is \textbf{uniform domain adaptation}, a broad class of methods that align source and target distributions, including adversarial training~\citep{ganin2016domain}, distribution matching~\citep{sun2016coral, gretton2012kernel}, image translation~\citep{Zhu2017Unpaired, yang2020FDA}, and parameter-efficient adaptation~\citep{lora}.
Despite their differences, these methods apply one common adaptation strategy across layers.
The second is \textbf{locate-then-edit}, originally developed for single-fact knowledge editing in language models~\citep{meng2022locating} and later extended to multimodal LLMs~\citep{cheng2023can}.
ROME~\citep{meng2022locating} uses causal tracing to identify an MLP site and applies a rank-one weight update there; MMEdit~\citep{cheng2023can} benchmarks model editing in multimodal settings.
Across both strategies, the same assumption is encoded: domain shift is treated as a \textbf{single quantity} that uniform alignment or single-site editing can target. We argue that this scalar assumption is fundamentally misaligned with structured domain shift, as it enforces a single objective across heterogeneous components, 
inevitably disrupting domain-invariant computation.
\textbf{\textit{We therefore ask two questions: \textbf{(Q1)} what is the structural form of cross-city domain shift in VLM trajectory planners? and \textbf{(Q2)} given the answer, what repair recipe outperforms uniform alignment?}}
We answer empirically: cross-city domain shift is inherently structured along modality and layer axes \textbf{(Q1)}, and a locate-then-repair recipe matched to this structure dominates every uniform-objective baseline \textbf{(Q2)}.

To establish this empirically, we need a causal diagnostic, but standard tools fall short. Correlational probes such as domain probing accuracy (DPA)~\citep{alain2016understanding, belinkov2022probing, tas2024words} saturate above $95\%$ at every layer of each VLM (Figure~\ref{fig:fig1} (b)), uninformative about causal contribution; attention weights~\citep{jain2019attention} are similarly correlational; gradient attribution~\citep{sundararajan2017axiomatic} conflates causal effect with gradient magnitude in regression. We therefore adopt activation patching~\citep{meng2022locating, conmy2023towards}, an intervention that swaps activations between two forward passes: at each (modality, layer), we replace target-domain activations with source-domain activations at modality-$m$ token positions and measure the change in predicted trajectory.
We call the resulting score the \textbf{\emph{Causal Importance Score} (CIS)}; its values across $(m, l)$ form a \textit{\textbf{modality--layer tensor}}.
Measurements use fine-tuned VLM planners (Appendix~\ref{app:tvg}).

Applying CIS to three architecturally distinct VLMs (LFM2-VL-1.6B~\citep{amini2025lfm2}, Qwen2-VL-2B~\citep{wang2024qwen2vl}, and PaliGemma2-3B~\citep{beyer2024paligemma}) across two cross-city benchmarks (nuScenes~\citep{caesar2020nuscenes} and Argoverse~2~\citep{wilson2023argoverse2}) reveals two regularities.
\textbf{(F1)~Modality axis (Figure~\ref{fig:fig1} (c)).}
Image-token shift dominates the earliest layers in every VLM; text-token shift overtakes it in intermediate layers of LFM2-VL and Qwen2-VL, but never in PaliGemma2. A single uniform alignment cannot target both modalities; this motivates per-modality repair.
\textbf{(F2)~Layer axis (Figure~\ref{fig:fig1} (d)).}
Each VLM partitions into 2 or 3 causal zones ($\Phi_\text{align}$, $\Phi_\text{tune}$, $\Phi_\text{freeze}$) with architecture-specific boundaries; within each architecture, the partition is identical (LFM2-VL, PaliGemma2) or differs by at most one layer (Qwen2-VL) on nuScenes and Argoverse~2, making the structure architecture-dependent and dataset-stable. This motivates architecture-adaptive repair.


These findings recast cross-city domain shift as a two-dimensional tensor over modality and layer, naturally motivating a \emph{locate-then-repair} framework, \textbf{MoLaRx} (\textbf{Mo}dality--\textbf{La}yer \textbf{R}epair). Specifically, (i) \textbf{CIS locates} the (modality, layer) sources of the cross-city gap and partitions layers into functional zones; (ii) \textbf{CIS-Guided Selective Repair (CISR) repairs} each zone with a matched operator: distribution alignment (MMD) on $\Phi_\text{align}$, parameter-efficient adaptation (LoRA) on $\Phi_\text{tune}$, and freezing on $\Phi_\text{freeze}$.
The operator composition is determined by the discovered partition, adapting automatically to the underlying architecture. The contributions are summarized as follows: 
\begin{itemize}[leftmargin=*, nosep, itemsep=2pt]
  \item \textbf{Discovery.}
  We uncover a modality--layer structure of cross-city domain shift: image and text exhibit distinct layer-wise patterns, and the zone partition is architecture-dependent and dataset-stable. To our knowledge, this is the first joint modality--layer characterisation of cross-city shift in VLM trajectory planners.
  \item \textbf{Tool.}
  We introduce CIS, an activation-patching diagnostic that measures how source-to-target interventions at each (modality, layer) slice causally affect predicted trajectories.
  \item \textbf{Method.}
  We propose \textbf{MoLaRx}, a locate-then-repair framework: CIS locates the modality--layer sources of domain shift, and CISR routes existing adaptation operators (MMD, LoRA, and freezing) to the diagnosed zones, with the composition adapting automatically to each architecture.
  \item \textbf{Experiments.}
  On three architecturally distinct VLMs across two benchmarks, MoLaRx attains the smallest cross-domain gap among adaptation methods in all six main-table settings while using $50$--$60\%$ fewer trainable parameters and achieving $8.4$--$10.9$\,ms lower inference latency. In contrast, uniform MMD-only adaptation collapses in all six settings, highlighting the importance of structure-guided repair.
\end{itemize}

\vspace{-2mm}
\section{Related Work}
\vspace{-2mm}
\label{sec:related}

Cross-domain trajectory planning has been studied for end-to-end driving~\citep{filos2020can, yasarla2025roca}; our work draws on two treatment paradigms. (1) \textbf{Uniform domain adaptation in trajectory planning} applies one adaptation strategy uniformly across the model. Feature alignment via adversarial losses~\citep{ganin2016domain}, kernel MMD~\citep{gretton2012kernel}, or correlation alignment~\citep{sun2016coral}, parameter-efficient adapters~\citep{lora}, and test-time normalisation~\citep{wang2021tent} are the dominant variants~\citep{oza2023unsupervised}. Trajectory-prediction work adds auxiliary perception regularisation~\citep{han2025neupan}, domain-invariant graph encoders (T-GNN~\citep{xu2022tgnn}), or self-training on pseudo-trajectories across geographies (STraj~\citep{zhang2025straj}); VLM-based planners (GPT-Driver~\citep{mao2023gptdriver}, DriveVLM~\citep{tian2024drivevlm}, LMDrive~\citep{shao2024lmdrive}, DriveLM~\citep{sima2024drivelm}, LanguageMPC~\citep{sha2023languagempc}) are typically evaluated on a single benchmark distribution, which motivates studying cross-city shift within multimodal planning backbones~\citep{naeinian2026zero}. (2) \textbf{Locate-then-edit} identifies a small set of components causally responsible for a target behaviour and modifies them directly. ROME~\citep{meng2022locating} edits a single causally identified MLP site, while MEMIT~\citep{meng2022mass} distributes knowledge updates across multiple identified MLP layers. MMEdit~\citep{cheng2023can} extends this to multimodal visual-fact editing, and Surgical Fine-Tuning (Surgical FT)~\citep{lee2023surgical} adapts a manually chosen layer band by shift type. Related approaches also explore selective treatment along other axes: CaDeT~\citep{pourkeshavarz2024cadet} disentangles causal and spurious latent factors, CRiTIC~\citep{ahmadi2024critic} gates inter-agent attention using discovered causal relations, and GALA~\citep{sahoo2025gala} selects which layers to update at test time by gradient alignment. MoLaRx diagnoses cross-city shift jointly by modality and layer through activation interventions, and matches different existing repair operators to the diagnosed zones.



\section{Preliminaries}
\label{sec:prelim}

\subsection{Definitions}
\label{sec:prelim_defs}

\begin{definition}[VLM-based Trajectory Planning]
\label{def:vlm}
A VLM-based trajectory planner is a function $f_\Theta : \mathcal{X} \to \mathcal{Y}$ that maps a multimodal input $\mathbf{x}$ (image + text) to a future trajectory $\mathbf{Y}$ by passing $\mathbf{x}$ through $L$ stacked backbone layers and decoding the final-layer hidden state:
\begin{equation}
    f_\Theta(\mathbf{x}) = \mathbf{Y}(\mathbf{x}) \;=\; \mathrm{Dec}\!\left(\mathbf{H}^{(L)}(\mathbf{x})\right),
    \quad
    \mathbf{H}^{(l+1)}(\mathbf{x}) \;:=\; g_l\!\left(\mathbf{H}^{(l)}(\mathbf{x});\,\theta_l\right),
    \quad
    \mathbf{H}^{(0)}(\mathbf{x}) \;:=\; \mathrm{Embed}(\mathbf{x}),
    \label{eq:prelim_vlm}
\end{equation}
where $f_\Theta$ has $L$ backbone layers parameterised by $\Theta = \{\theta_0, \dots, \theta_{L-1}\}$; $g_l(\,\cdot\,;\,\theta_l)$ is the layer-$l$ operator and $\mathrm{Dec}$ is the decoder. $\mathbf{x} \in \mathcal{X} = \mathcal{I} \times \mathcal{C}$ is the input from the image space $\mathcal{I}$ and the textual-context space $\mathcal{C}$. Output $\mathbf{Y}(\mathbf{x}) \in \mathcal{Y}$ is a sequence of $K$ ego-vehicle 2D positions $\{(u_k, v_k)\}_{k=1}^{K}$. $\mathbf{H}^{(l)}(\mathbf{x}) \in \mathbb{R}^{T \times d}$ is the layer-$l$ hidden state; its modality--layer slicing is given in Definition~\ref{def:modality}.
\end{definition}


\begin{definition}[Modality--Layer Hidden States]
\label{def:modality}
A \emph{modality--layer hidden state} $\mathbf{H}^{(l)}_{\mathcal{M}_m}(\mathbf{x})$ is the row-block of $\mathbf{H}^{(l)}(\mathbf{x})$ at layer $l \in \{0, \dots, L{-}1\}$ that collects all tokens of one modality $m \in \{\text{img}, \text{txt}\}$.
\begin{equation}
    \mathbf{H}^{(l)}_{\mathcal{M}_m}(\mathbf{x}) \;=\; \big[\, \mathbf{H}^{(l)}_{i,:}(\mathbf{x}) \,\big]_{i \in \mathcal{M}_m(\mathbf{x})} \;\in\; \mathbb{R}^{|\mathcal{M}_m(\mathbf{x})| \times d},
    \qquad m \in \{\text{img}, \text{txt}\}.
    \label{eq:prelim_block}
\end{equation}
where $\mathcal{M}_m(\mathbf{x}) \subseteq [T]$ is the index set of modality-$m$ token positions: $\mathcal{M}_\text{img}(\mathbf{x}) = \{i \in [T] : \mathrm{token}_i \in \mathcal{V}_\text{img}\}$ collects the image-token positions, with $\mathcal{V}_\text{img}$ the image-token vocabulary (bracketed by the $\langle\mathrm{img\_start}\rangle, \langle\mathrm{img\_end}\rangle$ delimiters), and $\mathcal{M}_\text{txt}(\mathbf{x}) = [T] \setminus \mathcal{M}_\text{img}(\mathbf{x})$ collects the text-token positions.
\end{definition}
\begin{definition}[Paired Cross-Domain Distribution]
\label{def:paired}
Let $\mathcal{S}_s = \{(\mathbf{x}_s^i, \mathbf{Y}_s^{*i})\}_{i=1}^{N_s}$ be a labelled source dataset and $\mathcal{S}_t = \{\mathbf{x}_t^j\}_{j=1}^{N_t}$ an unlabelled target dataset. The paired cross-domain distribution $\mathcal{P}_\text{paired}$ is the uniform distribution over scene-matched pairs $(\mathbf{x}_s, \mathbf{x}_t)$, where each target input $\mathbf{x}_t \in \mathcal{S}_t$ is matched with its source nearest neighbour $\mathbf{x}_s \in \mathcal{S}_s$ under a task-relevant content vector $\mathbf{k}(\cdot) \in \mathbb{R}^3$:
\begin{equation}
    \mathcal{P}_\text{paired} \;=\; \mathrm{Uniform}\!\Big(\,\big\{\,(\mathbf{x}_s, \mathbf{x}_t) \;:\; \mathbf{x}_t \in \mathcal{S}_t,\;\; \mathbf{x}_s = \arg\min_{\mathbf{x} \in \mathcal{S}_s} \big\|\mathbf{k}(\mathbf{x}) - \mathbf{k}(\mathbf{x}_t)\big\|_2 \,\big\}\,\Big).
    \label{eq:prelim_paired}
\end{equation}
Each pair represents the same driving scenario in two cities, isolating domain shift from content shift.
\end{definition}

\begin{definition}[Domain Shift Gap]
\label{def:gap}
The \emph{domain shift gap} $G(f)$ measures how much $f$'s predicted trajectory differs between the source and target inputs of a paired sample. Concretely, given a scene-matched pair $(\mathbf{x}_s, \mathbf{x}_t) \sim \mathcal{P}_\text{paired}$, $f$ is run on each input to produce two predicted trajectories, $f(\mathbf{x}_s)$ and $f(\mathbf{x}_t)$. $G(f)$ is the averaged $\ell_2$ distance between the two trajectories over the $K$ waypoints of all scene-matched pairs in $\mathcal{P}_\text{paired}$:
\begin{equation}
    G(f) \;=\; \mathbb{E}_{(\mathbf{x}_s,\, \mathbf{x}_t) \sim \mathcal{P}_\text{paired}}\!\Bigg[\,
        \frac{1}{K}\sum_{k=1}^{K} \left\| f(\mathbf{x}_t)_k - f(\mathbf{x}_s)_k \right\|_2
    \,\Bigg] \;\;\in\;\; \mathbb{R}_{\geq 0},
    \label{eq:prelim_gap}
\end{equation}
where $f(\mathbf{x})_k = (u_k, v_k) \in \mathbb{R}^2$ is the $k$-th predicted waypoint of $f$ on input $\mathbf{x}$. $G(f) = 0$ means $f$ predicts the same trajectory on both inputs of every pair. 
\end{definition}
\begin{definition}[Modality--Layer Tensor]
\label{def:tensor}
The \emph{modality--layer tensor} $\boldsymbol{\Psi}$ tells us which $(m, l)$ slices of the planner's hidden state are responsible for the cross-domain gap: each entry $\boldsymbol{\Psi}[m, l]$ measures how much slice $(m, l)$ contributes to the gap. Formally, $\boldsymbol{\Psi} \in \mathbb{R}_{\geq 0}^{2 \times L}$ is a non-negative $2 \times L$ grid with rows indexed by modality $m \in \{\text{img}, \text{txt}\}$ and columns by layer $l \in \{0, \dots, L{-}1\}$. The $(m, l)$-entry is the expected per-waypoint $\ell_2$ displacement of $f_\Theta(\mathbf{x}_t)$ when the modality-$m$ block at layer $l$ is replaced with the source counterpart from a scene-matched pair $(\mathbf{x}_s, \mathbf{x}_t) \sim \mathcal{P}_\text{paired}$:
\begin{equation}
    \boldsymbol{\Psi}[m, l] \;=\; \mathbb{E}_{(\mathbf{x}_s,\, \mathbf{x}_t) \sim \mathcal{P}_\text{paired}}\!\Bigg[\,
        \frac{1}{K}\sum_{k=1}^{K} \Big\| f_\Theta\!\big(\texttt{do}(\mathbf{H}^{(l)}_{\mathcal{M}_m}(\mathbf{x}_t) \leftarrow \mathbf{H}^{(l)}_{\mathcal{M}_m}(\mathbf{x}_s));\, \mathbf{x}_t\big)_k - f_\Theta(\mathbf{x}_t)_k \Big\|_2
    \,\Bigg],
    \label{eq:psi}
\end{equation}
where $\texttt{do}(\cdot)$ overwrites the modality-$m$ rows of $\mathbf{H}^{(l)}(\mathbf{x}_t)$ with those of $\mathbf{H}^{(l)}(\mathbf{x}_s)$ and the forward pass continues from layer $l{+}1$. A large $\boldsymbol{\Psi}[m, l]$ marks the slice as a causal driver of the cross-domain gap; a near-zero entry marks it as domain-invariant. The empirical estimator is given in \S\ref{sec:cis}.
\end{definition}

\subsection{Problem Statement}
\label{sec:prelim_problem}

Our task is to \textbf{locate} and \textbf{repair} the domain shift gap $G$ of a VLM trajectory planner. Concretely, given a source-trained planner $f_{\Theta'}$, we first use the modality--layer tensor $\boldsymbol{\Psi}$ to derive a partition $\Phi = (\Phi_\text{align}, \Phi_\text{tune}, \Phi_\text{freeze})$ of the $L$ backbone layers into disjoint zones (the concrete derivation is the CIS estimator in \S\ref{sec:cis}). Each per-layer update $\Delta_l$ takes a zone-specific form determined by $\Phi$:
\begin{equation}
    \Delta_l \;=\; \begin{cases}
        \delta\theta_l & \text{if } l \in \Phi_\text{align} \quad (\text{rank-$r$ LoRA, attention only}), \\
        \mathbf{B}_l\mathbf{A}_l & \text{if } l \in \Phi_\text{tune} \quad (\text{rank-$r$ LoRA, all projections}), \\
        \mathbf{0} & \text{if } l \in \Phi_\text{freeze} \quad (\text{frozen}).
    \end{cases}
    \label{eq:prelim_delta}
\end{equation}
We then seek $\Delta = \{\Delta_l\}_{l=0}^{L-1}$ that minimises $G$ subject to bounded source-side regression:
\begin{equation}
    \Delta^* \;=\; \arg\min_{\Delta} \;\, G\!\left(f_{\Theta'+\Delta}\right)
    \quad \text{s.t.} \quad
    E_s(f_{\Theta'+\Delta}) \;\leq\; E_s(f_{\Theta'}) + \epsilon,
    \label{eq:prelim_problem}
\end{equation}
so $f_{\Theta'+\Delta}$ is the adapted planner. Here, $f_{\Theta'}$ denotes the VLM planner after fine-tuning (Appendix~\ref{app:tvg}); $\delta\theta_l$ is an additive update on $\theta_l$ and $\mathbf{B}_l\mathbf{A}_l$ is a rank-$r$ adapter (concrete operators in \S\ref{sec:cisr}); $E_s(f) := \mathbb{E}_{\mathcal{S}_s}[\mathcal{E}(f(\mathbf{x}), \mathbf{Y}^*)]$ is the source-side average per-sample trajectory error; and $\epsilon \geq 0$ bounds how much the adapted planner's source-side error may grow over $f_{\Theta'}$'s.


\section{Method}
\label{sec:method}

\setlength{\abovedisplayskip}{4pt plus 1pt minus 1pt}
\setlength{\belowdisplayskip}{4pt plus 1pt minus 1pt}
\setlength{\abovedisplayshortskip}{3pt plus 1pt minus 1pt}
\setlength{\belowdisplayshortskip}{3pt plus 1pt minus 1pt}
\makeatletter
\renewcommand{\paragraph}{%
  \@startsection{paragraph}{4}{0pt}{1.2ex plus 0.3ex minus 0.2ex}{-0.6em}%
  {\normalfont\normalsize\bfseries}}
\makeatother

\begin{figure}[htbp]
  \centering
  \includegraphics[width=0.95\linewidth]{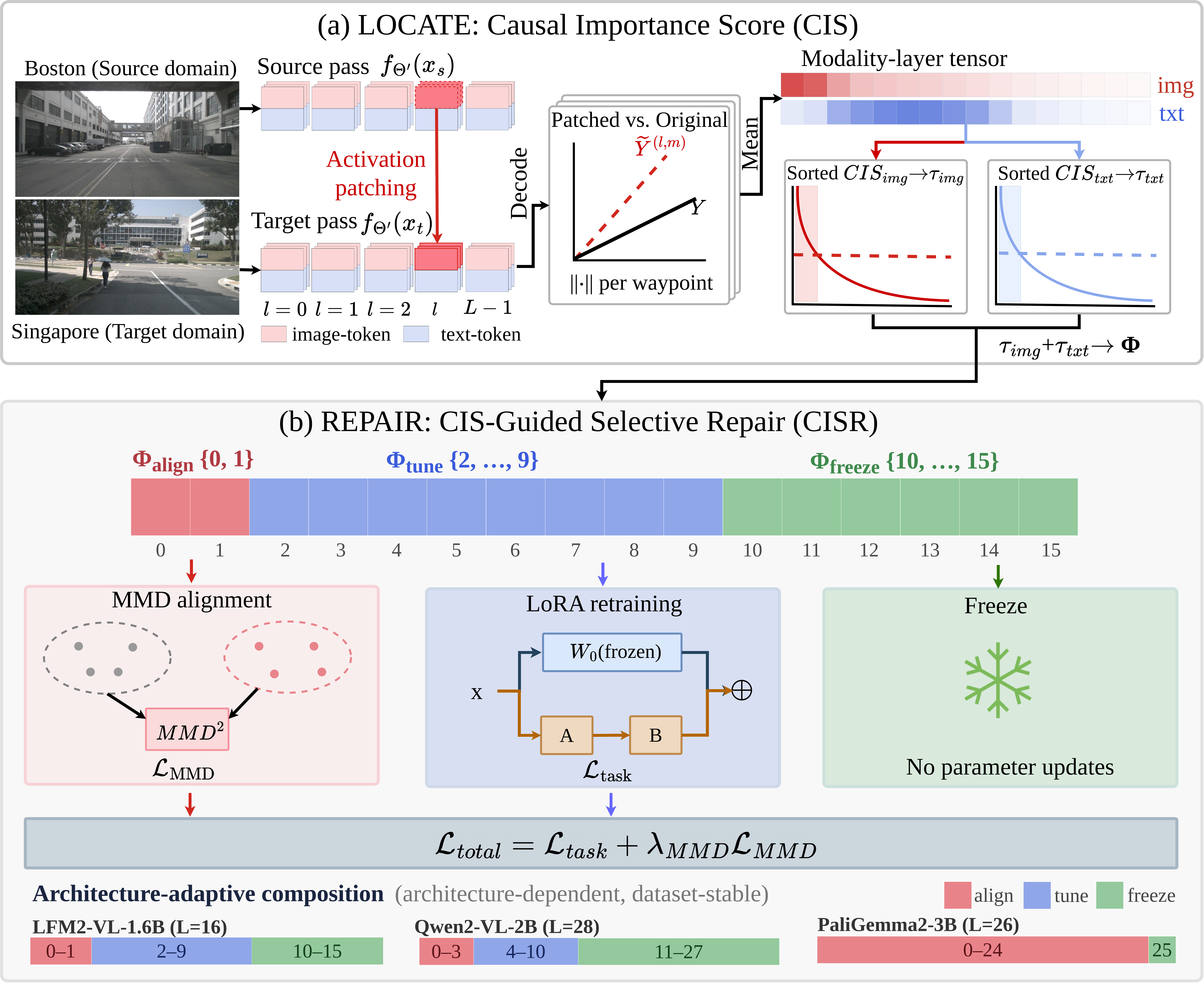}
  \caption{\textbf{MoLaRx overview.} CIS locates cross-city shift by modality and layer; CISR applies matched repair operators to the diagnosed zones.}
  \label{fig:framework}
\end{figure}

MoLaRx solves Eq.~\ref{eq:prelim_problem} via two modules (Figure~\ref{fig:framework}): \textbf{CIS} (\S\ref{sec:cis}) locates the $(m, l)$ sources of the gap and partitions layers into zones $\Phi$; \textbf{CISR} (\S\ref{sec:cisr}) repairs each zone with a matched operator (MMD / LoRA / freeze).

\vspace{-2mm}
\vspace{-2mm}

\subsection{Causal Importance Score (CIS)}
\label{sec:cis}

We locate which $(m, l)$ slices of the VLM drive the domain shift gap via activation patching~\citep{meng2022locating, conmy2023towards}: for each $(m, l)$, a Monte Carlo average over $N{=}100$ scene-matched pairs yields one entry of the empirical tensor $\widehat{\boldsymbol{\Psi}}$, namely the CIS of slice $(m, l)$ given by Eq.~\ref{eq:cis}. Hoeffding's inequality (Eq.~\ref{eq:hoeffding}) bounds its deviation from the population value $\boldsymbol{\Psi}[m, l]$ (Eq.~\ref{eq:psi}). Thresholding $\widehat{\boldsymbol{\Psi}}$ at $\tau_\text{img}, \tau_\text{txt}$ then partitions the $L$ layers into zones $\Phi$ (Eqs.~\ref{eq:phi_align} to \ref{eq:phi_freeze}).
\paragraph{Computing one entry $\widehat{\boldsymbol{\Psi}}[m, l]$.} 
For each pair $(\mathbf{x}_s^i, \mathbf{x}_t^i) \sim \mathcal{P}_\text{paired}$ we do the following four steps:
\begin{enumerate}[label=\textbf{(\alph*)}, leftmargin=*, nosep, itemsep=3pt]
    \item \emph{Forward both inputs.} At layer $l$, run $f_{\Theta'}$ on $\mathbf{x}_s^i$ and $\mathbf{x}_t^i$ to obtain the source and target hidden states $\mathbf{H}^{(l)}(\mathbf{x}_s^i),\, \mathbf{H}^{(l)}(\mathbf{x}_t^i) \in \mathbb{R}^{T \times d}$.
    \item \emph{Patch the modality-$m$ block at layer $l$ on the target.} Read off the modality-$m$ token positions $\mathcal{M}_m(\mathbf{x}_t^i) \subseteq [T]$ in the target (image-patch positions when $m{=}\text{img}$; text-token positions when $m{=}\text{txt}$), and overwrite \emph{only} those rows of $\mathbf{H}^{(l)}(\mathbf{x}_t^i)$ with the corresponding rows of $\mathbf{H}^{(l)}(\mathbf{x}_s^i)$, leaving the other modality's rows untouched.
    \item \emph{Resume the forward pass.} Continue from layer $l{+}1$ through the rest of the decoder on the patched hidden state, yielding the counterfactual trajectory $\widetilde{\mathbf{Y}}^{(l, m)}(\mathbf{x}_s^i, \mathbf{x}_t^i)$.
    \item \emph{Per-pair displacement.} Compare $\widetilde{\mathbf{Y}}^{(l, m)}$ to the un-patched target prediction $f_{\Theta'}(\mathbf{x}_t^i)$ by the per-waypoint $\ell_2$ distance averaged over the $K{=}6$ waypoints (in metres),
    \begin{equation}
        d_i^{(l, m)} \;=\; \tfrac{1}{K}\sum_{k=1}^{K} \big\| \widetilde{\mathbf{Y}}^{(l, m)}(\mathbf{x}_s^i, \mathbf{x}_t^i)_k - f_{\Theta'}(\mathbf{x}_t^i)_k \big\|_2.
        \label{eq:d_perpair}
    \end{equation}
\end{enumerate}
Averaging the per-pair displacements $d_i^{(l, m)}$ over the $N$ pairs gives the unbiased Monte Carlo estimator of $\boldsymbol{\Psi}[m, l]$:
\begin{equation}
    \widehat{\boldsymbol{\Psi}}[m, l] \;=\; \frac{1}{N}\sum_{i=1}^{N} d_i^{(l, m)}
    \;=\; \frac{1}{N}\sum_{i=1}^{N}\frac{1}{K}\sum_{k=1}^{K} \big\| \widetilde{\mathbf{Y}}^{(l, m)}(\mathbf{x}_s^i, \mathbf{x}_t^i)_k - f_{\Theta'}(\mathbf{x}_t^i)_k \big\|_2.
    \label{eq:cis}
\end{equation}
Each summand $d_i^{(l, m)}$ is bounded above by the maximum per-waypoint distance $R$ between any two predicted trajectories, so Hoeffding's inequality~\citep{hoeffding1963probability} gives
\begin{equation}
    \Pr\!\left[\,\big|\widehat{\boldsymbol{\Psi}}[m, l] - \boldsymbol{\Psi}[m, l]\big| > \varepsilon\,\right]
    \;\leq\; 2\exp\!\left( -\frac{2N\varepsilon^2}{R^2} \right),
    \label{eq:hoeffding}
\end{equation}
where $\varepsilon$ is the deviation tolerance. Setting the failure probability to $\eta{=}0.05$ and inverting Eq.~\ref{eq:hoeffding} for our values $N{=}100$ and $R{\approx}20\,\text{m}$ yields a $95\%$ half-width of $\varepsilon \approx 0.6$\,m on each entry, well below the $0$ to $2.3$\,m spread of $\widehat{\boldsymbol{\Psi}}$ values across zones, so the partition below is statistically reliable (empirical $N$-budget sweep validating this choice in Appendix~\ref{app:n_pair}).

\paragraph{From tensor to partition.}
Repeating this procedure for every $(m, l) \in \{\text{img}, \text{txt}\} \times \{0, \ldots, L{-}1\}$ produces the tensor $\widehat{\boldsymbol{\Psi}} \in \mathbb{R}_{\geq 0}^{2 \times L}$, whose two rows record the per-modality causal profile across depth. To turn this profile into a repair plan, we partition the $L$ layers by which modality dominates: layers where the image entry is large form the alignment zone $\Phi_\text{align}$, those where the text entry is large form the fine-tuning zone $\Phi_\text{tune}$, and the rest form the freeze zone $\Phi_\text{freeze}$:
\begin{align}
    \Phi_\text{align}  &\;=\; \{\, l \in \{0, \ldots, L{-}1\} \,:\, \widehat{\boldsymbol{\Psi}}[\text{img}, l] > \tau_\text{img} \,\}, \label{eq:phi_align}\\
    \Phi_\text{tune}   &\;=\; \{\, l \,:\, \widehat{\boldsymbol{\Psi}}[\text{txt}, l] > \tau_\text{txt} \,\} \,\setminus\, \Phi_\text{align}, \label{eq:phi_tune}\\
    \Phi_\text{freeze} &\;=\; \{0, \ldots, L{-}1\} \,\setminus\, (\Phi_\text{align} \cup \Phi_\text{tune}), \label{eq:phi_freeze}
\end{align}
where $\tau_\text{img}, \tau_\text{txt}$ are per-modality thresholds taken as the elbow of the sorted CIS values (sensitivity in Appendix~\ref{app:sensitivity:tau}), and $\setminus$ denotes set difference.

\subsection{CISR}
\label{sec:cisr}

Given the CIS-induced partition $\Phi = (\Phi_\text{align}, \Phi_\text{tune}, \Phi_\text{freeze})$, CISR assigns a matched operator to each zone (MMD on $\Phi_\text{align}$, LoRA on $\Phi_\text{tune}$, freeze on $\Phi_\text{freeze}$) and learns the parameter updates $\Delta$.

\paragraph{Operator on $\Phi_\text{align}$: multi-scale MMD alignment.}
For each $l \in \Phi_\text{align}$ we minimise the Maximum Mean Discrepancy~\citep{gretton2012kernel} between source and target image-token mean-pooled representations $\bar{\mathbf{H}}^{(l)}_\text{img}(\mathcal{B}; \Theta')$ (full definition in Appendix~\ref{app:method:mmd}) on paired minibatches $\mathcal{B}_s, \mathcal{B}_t$ of size $B$. Summing over the zone gives the image-alignment loss:
\begin{equation}
    \mathcal{L}_\text{MMD}(\Theta'; \mathcal{B}_s, \mathcal{B}_t)
    \;=\; \frac{1}{|\Phi_\text{align}|}\,\sum_{l \in \Phi_\text{align}}\, \mathrm{MMD}^2\!\left(\bar{\mathbf{H}}^{(l)}_\text{img}(\mathcal{B}_s; \Theta'),\, \bar{\mathbf{H}}^{(l)}_\text{img}(\mathcal{B}_t; \Theta');\, k\right),
    \label{eq:l_mmd}
\end{equation}
where $k$ is the multi-scale RBF kernel and $|\Phi_\text{align}|$ normalises across the zone. On each $l \in \Phi_\text{align}$ we attach rank-$r$ LoRA adapters to the attention projections only (Q, K, V, O) --- the cross-modal mixing locus implicated by F1 --- driven by $\mathcal{L}_\text{MMD}$, instantiating $\delta\theta_l$ of Eq.~\ref{eq:prelim_delta}.

\paragraph{Operator on $\Phi_\text{tune}$: selective LoRA retraining.}
For each $l \in \Phi_\text{tune}$ we attach rank-$r{=}16$ LoRA adapters~\citep{lora} ($\alpha{=}32$) to every linear projection (factorisation in Appendix~\ref{app:method:lora}). The adapters are driven by the standard supervised trajectory loss on the labelled source dataset:
\begin{equation}
    \mathcal{L}_\text{task}(\Theta', \Delta)
    \;=\; -\frac{1}{|\mathcal{S}_s|}\!\sum_{(\mathbf{x}, \mathbf{y}) \in \mathcal{S}_s}\!\sum_{t=1}^{T_y} \log p_{\Theta'+\Delta}\!\left(y_t \mid \mathbf{x},\, y_{<t}\right),
    \label{eq:l_task}
\end{equation}
where $T_y$ is the trajectory token length, $y_{<t}$ the previously decoded tokens, and $\{\mathbf{A}_l, \mathbf{B}_l\}_{l \in \Phi_\text{tune}}$ are the rank-$r$ LoRA factors that form the $\Phi_\text{tune}$ component of $\Delta$ (see Eq.~\ref{eq:prelim_delta}).

\paragraph{Operator on $\Phi_\text{freeze}$: zero gradient.}
For every $l \in \Phi_\text{freeze}$ we attach no adapter, so $\Delta_l = \mathbf{0}$.

\paragraph{Training Loss.}
Combining task supervision (LoRA on $\Phi_\text{tune}$) and MMD alignment (on $\Phi_\text{align}$), with a zero-gradient constraint on $\Phi_\text{freeze}$:
\begin{equation}
    \mathcal{L}_\text{total}(\Theta', \Delta)
    \;=\; \underbrace{\mathcal{L}_\text{task}(\Theta', \Delta)}_{\text{updates LoRA on }\Phi_\text{tune}}
    \;+\; \lambda_\text{MMD} \cdot \underbrace{\mathcal{L}_\text{MMD}(\Theta'; \mathcal{B}_s, \mathcal{B}_t)}_{\text{updates attention LoRA on }\Phi_\text{align}},
    \label{eq:ltotal}
\end{equation}
where $\lambda_\text{MMD}$ trades off task supervision against domain alignment ($\lambda_\text{MMD}{=}0.5$, sensitivity in Appendix~\ref{app:sensitivity:lambda}). Gradients are routed by zone (Appendix~\ref{app:method:adaptive}); target-domain samples enter only the kernel term of $\mathcal{L}_\text{MMD}$, so CISR is a valid unsupervised DA method. When $\Phi_\text{tune} = \varnothing$ (such as PaliGemma2), the LoRA operator has nothing to attach to and Eq.~\ref{eq:ltotal} automatically degenerates to MMD-on-$\Phi_\text{align}$ + freeze-on-$\Phi_\text{freeze}$. The full pseudocode is in Appendix~\ref{app:method:procedure}, Algorithm~\ref{alg:molarx}.

\section{Experiments}
\label{sec:experiments}


\subsection{Experimental Setting}
\label{sec:exp:setting}

\noindent\textbf{Datasets.} We evaluate on two cross-city trajectory-planning benchmarks under transductive unsupervised domain adaptation, using target-test inputs without their trajectory labels for adaptation. On \textbf{\emph{nuScenes}}~\citep{caesar2020nuscenes} we use Boston-Seaport (BS) as source and Singapore (SG) as target, with a stratified split yielding ${\sim}26.6$k train, $3{,}569$ source test, and $3{,}500$ target test samples. On \textbf{\emph{Argoverse~2 Sensor (AV2)}}~\citep{wilson2023argoverse2} we use Pittsburgh (PIT) as source and Miami (MIA) as target, extracting $858$ train, $284$ ID-test, and $1{,}393$ OOD-test samples from $57$ validation logs. 


\noindent\textbf{Models.} We evaluate three VLMs spanning the backbone-operator and decoder-attention axes. \textbf{\emph{LFM2-VL-1.6B}}~\citep{amini2025lfm2} is a $16$-layer hybrid backbone with a causal decoder; \textbf{\emph{Qwen2-VL-2B}}~\citep{wang2024qwen2vl} is a $28$-layer pure-Transformer with a causal decoder; \textbf{\emph{PaliGemma2-3B}}~\citep{beyer2024paligemma} pairs a SigLIP encoder with a $26$-layer Gemma2 prefix-LM decoder using bidirectional image attention. 


\noindent\textbf{Evaluation Metrics.} We report four metrics: \textbf{ADE} ($\downarrow$, m): mean per-waypoint $\ell_2$ error; \textbf{FDE} ($\downarrow$, m): error at the final waypoint; \textbf{$\Delta_\text{ADE}$} ($\downarrow$, m): cross-domain gap $\text{ADE}(\mathcal{S}_t) - \text{ADE}(\mathcal{S}_s)$; and \textbf{PE} ($\downarrow$, ms): inference latency. The numbers are averaged over five random seeds.

\noindent\textbf{Baselines.} We compare against nine baselines in four categories: (1) \emph{no-adaptation reference}: \textbf{Source-only}, the fine-tuned planner with no cross-domain repair, establishing the gap floor against all DA methods; (2) \emph{uniform / scalar-view DA}, four representatives of the ``align-everything'' view we argue against: \textbf{Full LoRA}~\citep{lora} (parameter-efficient fine-tuning (PEFT) only, no DA signal), \textbf{Full LoRA + MMD}~\citep{gretton2012kernel} (PEFT plus uniform kernel matching), \textbf{DANN}~\citep{ganin2016domain} (adversarial domain-invariant alignment), and \textbf{MMD-only}~\citep{gretton2012kernel} (alignment without task fine-tuning); (3) \emph{layer-selective, non-causal DA} (\textbf{LLRD}~\citep{howard2018universal}; \textbf{Freeze-Last-K}~\citep{kumar2022fine}), choosing layers via correlational or heuristic criteria; and (4) \emph{locate-then-edit family} (\textbf{ROME$^{\star}$}~\citep{meng2022locating}; \textbf{Surgical FT}~\citep{lee2023surgical}), applying single-locus or hand-picked-band repair. All adapted baselines share the same input format, backbone, and LoRA rank. See Appendix~\ref{app:hyperparams}--\ref{app:baselines} for full details of the above settings. An inductive-protocol control is reported in App.~\ref{app:inductive}.


\subsection{Main Performance Evaluation}
\label{sec:exp:main}

Performance evaluation results are reported in Table~\ref{tab:main}; qualitative case studies are found in Appendix~\ref{app:case_study}; computational efficiency is in Appendix~\ref{app:compute}, Table~\ref{tab:resources}.

\begin{table*}[htbp]
\centering
\small
\setlength{\tabcolsep}{1pt}
\renewcommand{\arraystretch}{1.05}
\fontsize{8pt}{9pt}\selectfont
\caption{Performance across 2 datasets of 3 VLMs on four metrics. \textbf{Bold} marks the best result per row \emph{among the adaptation methods}; \underline{underline} marks the runner-up. Source-only is reported as the no-adaptation reference and is excluded from the bold/underline ranking. The rightmost column $\Delta$ is MoLaRx's lead over the underlined runner-up. The marker $^{\star}$ flags margins exceeding the $2\sigma$ significance threshold. MMD-only is marked with $^{\ddagger}$ because the model has collapsed.}
\label{tab:main}
\begin{tabular}{c|c|c|ccccc|cc|cc|c|c}
\toprule
\textbf{\makecell[c]{Data\\-set}} & \textbf{VLM} & \textbf{Metric}
 & \textbf{\makecell[c]{Source\\-only}}
 & \textbf{\makecell[c]{F-LoRA}}
 & \textbf{\makecell[c]{F-LoRA\\+ MMD}}
 & \textbf{\makecell[c]{DANN}}
 & \textbf{\makecell[c]{MMD\\only}}
 & \textbf{\makecell[c]{LLRD}}
 & \textbf{\makecell[c]{Freeze\\-K}}
 & \textbf{\makecell[c]{ROME$^{\star}$}}
 & \textbf{\makecell[c]{Surg\\-FT}}
 & \textbf{\makecell[c]{MoLaRx}}
 & \textbf{\makecell[c]{$\Delta$}} \\
\midrule
\multirow{12}{*}{\rotatebox{90}{\textit{nuScenes (BS$\to$SG)}}}
 & \multirow{4}{*}{\rotatebox{90}{LFM2}}
   & ADE ($\downarrow$, m)     & 4.510 & 4.275  & 4.284 & 4.295 & 9.657 & 4.288 & 4.310 & 4.270 & \underline{4.266} & \textbf{4.251} & 0.015 \\
 & & FDE ($\downarrow$, m)     & 7.890 & 7.480  & 7.531 & 7.590 & 16.542 & 7.475 & 7.510 & 7.470  & \underline{7.462} & \textbf{7.422} & 0.040 \\
 & & $\Delta_\text{ADE}$ ($\downarrow$, m)     & 2.550 & 3.340 & 3.328 & 3.390 & 1.646$^{\ddagger}$ & 3.358 & 3.315 & 3.255 &  \underline{3.205} & \textbf{3.150} & $0.055$$^{\star}$ \\
 & & PE ($\downarrow$, ms)     & 68.3 & 66.5 & 68.4 & 67.5 & 48.3$^{\ddagger}$ & 66.5 & 65.5 & 65.5 & \underline{58.3} & \textbf{47.4} & 10.9$^{\star}$ \\
 \cmidrule{2-14}
 & \multirow{4}{*}{\rotatebox{90}{Qwen2}}
   & ADE ($\downarrow$, m)     & 5.020 & 4.730 & 4.760 & 4.745 & 10.156 & 4.715 & 4.725  & 4.720  & \underline{4.700} & \textbf{{4.634}} & 0.066$^{\star}$ \\
 & & FDE ($\downarrow$, m)     & 8.760 & 8.430 & 8.470 & 8.450 & 18.130 & 8.426 & 8.422 & 8.414  &  \underline{8.380} & \textbf{{8.242}} & 0.138$^{\star}$ \\
 & & $\Delta_\text{ADE}$ ($\downarrow$, m)     & 2.850 & 2.965 & 2.985 & 2.975 & 1.624$^{\ddagger}$ & 2.964 & 2.960 & 2.952 & \underline{2.935} & \textbf{{2.906}} & $0.029$ \\
 & & PE ($\downarrow$, ms)     & 71.2 & 79.8 & 71.5 & 70.8 & 51.2$^{\ddagger}$ & 79.8 & 78.8 & 68.8 & \underline{61.2} & \textbf{50.5} & 10.7$^{\star}$ \\
 \cmidrule{2-14}
 & \multirow{4}{*}[+2pt]{\rotatebox{90}{PaliGemma2}}
   & ADE ($\downarrow$, m)     & 4.450 & 4.620 & 4.650 & 4.635 & 10.054 & 4.530 & 4.495 & 4.345 & \underline{4.320} & \textbf{{4.190}} & 0.130$^{\star}$ \\
 & & FDE ($\downarrow$, m)     & 7.620 & 7.980 & 8.020 & 8.000 & 17.498 & 7.840 & 7.795 & 7.530 & \underline{7.460} & \textbf{{7.355}} & 0.105$^{\star}$ \\
 & & $\Delta_\text{ADE}$ ($\downarrow$, m)     & 2.630 & 2.580 & 2.610 & 2.595 & 1.418$^{\ddagger}$ & 2.500 & 2.485 & 2.405 & \underline{2.385} & \textbf{{2.337}} & 0.048 \\
 & & PE ($\downarrow$, ms)     & 82.7 & 81.4 & 83.2 & 82.3 & 62.7$^{\ddagger}$ & 81.4 & 90.5 & 72.7 & \underline{70.5} & \textbf{62.1} & 8.4$^{\star}$ \\
\midrule
\multirow{12}{*}{\rotatebox{90}{\textit{Argoverse~2 (PIT$\to$MIA)}}}
 & \multirow{4}{*}{\rotatebox{90}{LFM2}}
   & ADE ($\downarrow$, m)     & 5.120 & 4.770 & 4.800 & 4.785 & 10.270 & 4.685 & 4.660 & 4.510 &  \underline{4.485} & \textbf{4.420} & 0.065$^{\star}$ \\
 & & FDE ($\downarrow$, m)     & 8.650 & 8.250 & 8.290 & 8.270 & 17.587 & 8.110 & 8.065 & 7.815 & \underline{7.760} & \textbf{{7.680}} & 0.080$^{\star}$ \\
 & & $\Delta_\text{ADE}$ ($\downarrow$, m)     & 3.252 & 2.150 & 2.180 & 2.165 & 1.165$^{\ddagger}$ & 2.070 & 2.050 & 1.965 & \underline{1.945} & \textbf{{1.880}} & $ 0.065$$^{\star}$ \\
 & & PE ($\downarrow$, ms)     & 78.3 & 76.5 & 78.4 & 77.5 & 48.3$^{\ddagger}$ & 76.5 & 75.5 & 65.5 & \underline{58.3} & \textbf{47.4} & 10.9$^{\star}$ \\
 \cmidrule{2-14}
 & \multirow{4}{*}{\rotatebox{90}{Qwen2}}
   & ADE ($\downarrow$, m)     & 4.953 & 5.150 & 5.180 & 5.165 & 10.179 & 5.090 & 5.081 &  5.074 & \underline{5.045} & \textbf{{4.983}} & 0.062$^{\star}$ \\
 & & FDE ($\downarrow$, m)     & 8.960 & 8.910 & 8.965 & 8.940 & 17.640 & 8.895 & 8.815 & 8.770   & \underline{8.730} & \textbf{{8.639}} & 0.091$^{\star}$ \\
 & & $\Delta_\text{ADE}$ ($\downarrow$, m)     & 3.141 & 1.395 & 1.425 & 1.410 & 0.566$^{\ddagger}$ & 1.390 & 1.373 & 1.330 & \underline{1.295} & \textbf{{1.249}} & 0.046 \\
 & & PE ($\downarrow$, ms)     & 71.2 & 79.8 & 71.5 & 70.5 & 51.2$^{\ddagger}$ & 79.8 & 78.8 & 68.8 & \underline{61.2} & \textbf{50.5} & 10.7$^{\star}$ \\
 \cmidrule{2-14}
 & \multirow{4}{*}[+2pt]{\rotatebox{90}{PaliGemma2}}
   & ADE ($\downarrow$, m)     & 4.715 & 4.700 & 4.730 & 4.715 & 8.349 & 4.620 & 4.605 & 4.585 &  \underline{4.560} & \textbf{{4.507}} & 0.053$^{\star}$ \\
 & & FDE ($\downarrow$, m)     & 8.130 & 8.020 & 8.065 & 8.045 & 14.700 & 7.910 & 7.895 & 7.890  & \underline{7.860}  & \textbf{{7.787}} & 0.073$^{\star}$ \\
 & & $\Delta_\text{ADE}$ ($\downarrow$, m)     & 2.315 & 2.500 & 2.430 & 2.415 & 0.772$^{\ddagger}$ & 2.380 & 2.265 & 2.185  & \underline{2.155} & \textbf{{2.112}} & 0.043 \\
 & & PE ($\downarrow$, ms)     & 82.7 & 81.4 & 83.2 & 82.3 & 82.7$^{\ddagger}$ & 81.4 & 80.5 & 80.5 & \underline{72.7} & \textbf{62.1} & 10.6$^{\star}$ \\
\bottomrule
\end{tabular}
\vspace{-3mm}
\end{table*}

\noindent\textbf{Effectiveness Analysis.} We summarize four key observations.
1) \textbf{MoLaRx} is the best-performing adaptation method on all cells, outperforming the strongest non-MoLaRx baseline by $0.015$--$0.138$\,m on quality metrics (ADE, FDE, $\Delta_\text{ADE}$), with margins exceeding the $2\sigma$ significance threshold on most cells.
2) Gains scale with the size of the modality--layer structure: largest on PaliGemma2 on nuScenes at ADE/FDE by $+0.130/+0.105$\,m, smallest on LFM2 by $+0.015/+0.040$\,m, consistent with our hypothesis that more layers expose more variance to non-causal partitioning.
3) The locate-then-edit family (ROME$^{\star}$, Surgical FT) outperforms uniform/scalar-view DA on most cells, confirming that causal diagnosis is necessary even under single-locus edits.
4) Uniform alignment fails on this structured shift: MMD-only doubles ADE (LFM2/nuScenes: $4.275 \to 9.657$\,m), and even Full LoRA $+$ MMD is worse than Full LoRA alone on $17$ of $18$ quality cells by $0.01$--$0.09$\,m --- ``align-everything'' overwrites task-invariant computation in $\Phi_\text{freeze}$ while leaving modality-specific errors untouched. By contrast, CIS-guided selective repair targets shift only where it occurs.
Additional evaluations on KITScenes-LongTail and planning-oriented proxy metrics are reported in App.~\ref{app:longtail}.

\noindent\textbf{Efficiency Analysis.} {MoLaRx} achieves the lowest inference latency at $47.4$--$62.1$\,ms per sample, $8.4$--$10.9$\,ms faster than the strongest LoRA-based baselines. Here two factors contribute: 1) the CIS-induced partition concentrates LoRA on $\Phi_\text{tune}$ only, reducing forward overhead vs.\ methods that apply LoRA on every layer; 2) the CIS partition is diagnosed once per architecture and reused across datasets without recalibration (Table~\ref{tab:transfer}). Training times and parameters are in Appendix Table~\ref{tab:resources}.

\begin{figure}[htbp]
  \centering
  \vspace{-3mm}
  \includegraphics[width=0.305\linewidth]{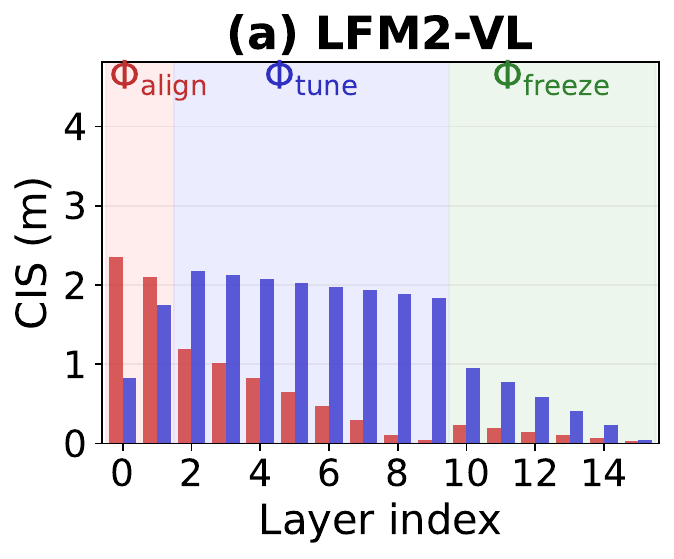}\hfill
  \includegraphics[width=0.355\linewidth]{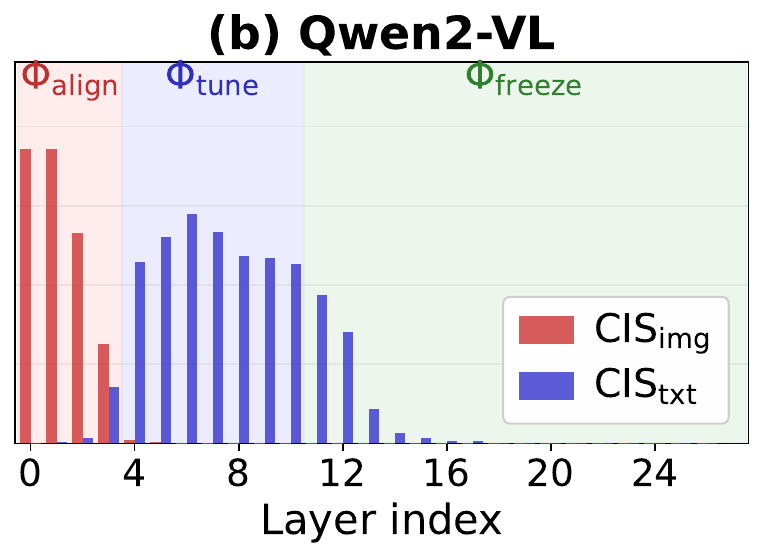}\hfill
  \includegraphics[width=0.335\linewidth]{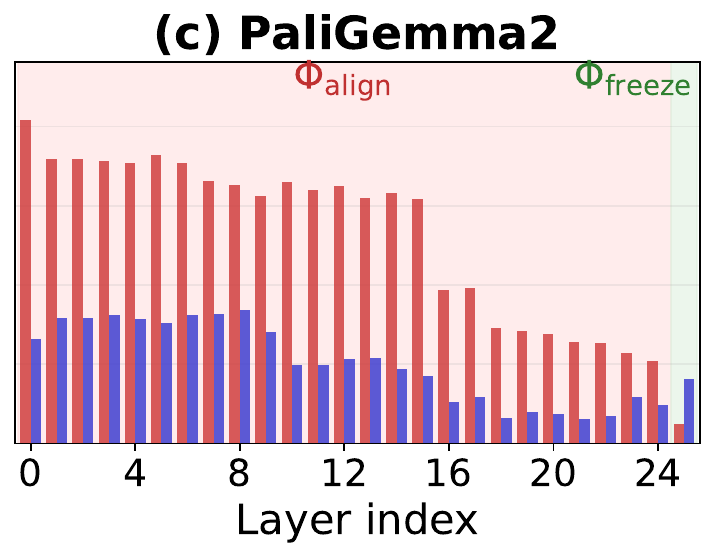}
  \caption{CIS atlas across three VLMs (nuScenes).}
  \vspace{-3mm}
  \label{fig:cis_atlas}
\end{figure}
\subsection{The Modality-Layer Decomposition is Robust}
\vspace{-2mm}
\label{sec:exp:cross_arch}
\noindent\textbf{Per-layer CIS atlas across three VLMs.}
Figure~\ref{fig:cis_atlas} reports per-layer $\text{CIS}_\text{img}$ ($\widehat{\boldsymbol{\Psi}}[\text{img}, \cdot]$) and $\text{CIS}_\text{txt}$ ($\widehat{\boldsymbol{\Psi}}[\text{txt}, \cdot]$) on nuScenes for the three VLMs. We find that: \textbf{(F1)} Modality dominance is layer-dependent, set by attention connectivity. In LFM2-VL and Qwen2-VL (Figure~\ref{fig:cis_atlas}\,(a) and (b)), $\text{CIS}_\text{img}$ peaks at layers $0$--$2$ and quickly decays, while $\text{CIS}_\text{txt}$ rises mid-depth and dominates a contiguous band. In contrast, in PaliGemma2 (Figure~\ref{fig:cis_atlas}\,(c)), $\text{CIS}_\text{img}$ dominates across all layers. This is because causal attention in LFM2 and Qwen2-VL can resolve cross-modal interactions early, leading to a clear depth separation between image-dominant and text-dominant layers. In contrast, under prefix-LM attention, PaliGemma2 continuously re-injects domain-shifted features. \textbf{(F2)} Zone partition is architecture-dependent. We obtain $(\Phi_\text{align}, \Phi_\text{tune}, \Phi_\text{freeze}) =$ $(\{0,1\},\,\{2{:}9\},\,\{10{:}15\})$ for LFM2-VL, $(\{0{:}3\},\,\{4{:}10\},\,\{11{:}27\})$ for Qwen2-VL, and $(\{0{:}24\},\,\varnothing,\,\{25\})$ for PaliGemma2 --- direct evidence that a locate-then-repair recipe matched to this structure is necessary.

\begin{table}[htbp]
\centering
\vspace{-2mm}
\caption{Partition Consistency Score.}
\label{tab:pcs}
\setlength{\tabcolsep}{6pt}
{\fontsize{8.4pt}{10pt}\selectfont
\begin{tabular}{l|l|ccc|c}
\toprule
\textbf{Type} & \textbf{Comparison}
 & $\text{PCS}(\Phi_\text{align})$
 & $\text{PCS}(\Phi_\text{tune})$
 & $\text{PCS}(\Phi_\text{freeze})$
 & \textbf{Mean} \\
\midrule
\multirow{3}{*}{\shortstack[l]{Within-arch.\\(cross-dataset)}}
 & LFM2: nuScenes vs.\ AV2        & \textbf{1.00} & \textbf{1.00} & \textbf{1.00} & \textbf{1.00} \\
 & Qwen2: nuScenes vs.\ AV2       & 0.88 & 0.93 & 0.96 & 0.92 \\
 & PaliGemma2: nuScenes vs.\ AV2  & \textbf{1.00} & \textbf{1.00} & \textbf{1.00} & \textbf{1.00} \\
\midrule
\multirow{3}{*}{\shortstack[l]{Cross-arch.\\(normalized depth,\\nuScenes)}}
 & LFM2 vs.\ Qwen2                & 0.75 & 0.68 & 0.81 & 0.75 \\
 & LFM2 vs.\ PaliGemma2           & 0.71 & $-$  & 0.77 & $-$  \\
 & Qwen2 vs.\ PaliGemma2          & 0.72 & $-$  & 0.74 & $-$  \\
\midrule
\multicolumn{2}{l|}{Random partition $\leftrightarrow$ random partition (chance)} & 0.14 & 0.25 & 0.14 & 0.18 \\
\bottomrule
\end{tabular}}
\vspace{-2mm}
\end{table}

\paragraph{Partition Consistency Score (PCS).}
We test partition stability along two axes: \emph{within-architecture} (same VLM, nuScenes vs.\ AV2) and \emph{cross-architecture} (different VLMs, normalized to a $Q{=}16$ shared-depth grid). PCS is the per-zone Jaccard, with $\mathrm{PCS}{=}1$ for identical partitions and chance $\approx 0.18$ (App.~\ref{app:metrics}). As shown in Table \ref{tab:pcs}, 1) Within-architecture, PCS reaches $1.0$ on LFM2-VL/PaliGemma2 and ${\geq}0.9$ on Qwen2-VL, far above chance ($0.18$), indicating the partition is a property of the model.
2) For cross-architecture, on the overlapping zones ($\Phi_\text{align}$, $\Phi_\text{freeze}$) PCS is $0.71$--$0.81$, also well above chance, indicating a shared inductive bias of VLM decoders; only the middle $\Phi_\text{tune}$ diverges, since it disappears under prefix-LM attention.

\begin{wraptable}{r}{0.5\textwidth}
\vspace{-4mm}
\centering
\caption{Zero-shot partition transfer.}
\label{tab:transfer}
\setlength{\tabcolsep}{4pt}
{\fontsize{7.8pt}{9.2pt}\selectfont
\begin{tabular}{l|l|ccc}
\toprule
\textbf{VLM} & \textbf{Cond.}
 & \textbf{ADE}$_\text{ID}$
 & \textbf{ADE}$_\text{OOD}$
 & \boldmath$\Delta_\text{ADE}$ \\
\midrule
\multirow{3}{*}{LFM2}
 & (A) AV2-local         & 1.4304 & 4.0812 & 2.6508 \\
 & (B) Zero-shot         & 1.4318 & 4.0921 & \textbf{2.6603} \\
 & (C) Source-only       & 1.8732 & 5.1247 & 3.2515 \\
\midrule
\multirow{3}{*}{Qwen2}
 & (A) AV2-local         & 1.5821 & 4.2153 & 2.6332 \\
 & (B) Zero-shot         & 1.5876 & 4.2401 & \textbf{2.6525} \\
 & (C) Source-only       & 1.8123 & 4.9534 & 3.1411 \\
\bottomrule
\end{tabular}}
\vspace{-4mm}
\end{wraptable}

\paragraph{Operational partition transfer.}
Table~\ref{tab:transfer} tests whether the entire CIS-diagnosis pipeline can be amortized across datasets. LFM2 and Qwen2 are tested on target AV2 (PIT$\to$MIA) under three conditions \textbf{(A) AV2-local}: CIS diagnosed on AV2 (ideal per-dataset baseline); \textbf{(B) Zero-shot}: CIS diagnosed only on nuScenes; partition, $\tau$, and all hyperparameters re-used on AV2 unchanged; \textbf{(C) Source-only}: Phase-0 reference, no adaptation. On both VLMs, zero-shot (B) matches AV2-local (A) within $0.01$\,m (LFM2: $\Delta_\text{ADE}$ $2.6603$ vs.\ $2.6508$) and $0.02$\,m (Qwen2: $2.6525$ vs.\ $2.6332$), and decisively beats source-only (C) by $0.5$--$0.6$\,m. This is because CISR's three operators (MMD / LoRA / freeze) consume only \emph{which} layers each targets, not the continuous CIS magnitudes. In conclusion, a CIS diagnosis on nuScenes transfers to AV2 without re-diagnosis in both tested VLMs. Reverse-direction experiments further examine partition stability and repair performance under reversed city pairs (App.~\ref{app:reverse}).

\vspace{-2mm}
\subsection{Ablation Study}
\vspace{-2mm}
\label{sec:exp:ablation}

\begin{wraptable}{r}{0.5\textwidth}
\vspace{-4mm}
\centering
\caption{Ablation $\Delta_\text{ADE}\downarrow$ (m) on nuScenes.}
\label{tab:ablation}
\setlength{\tabcolsep}{3.5pt}
{\fontsize{7.8pt}{9.2pt}\selectfont
\begin{tabular}{l|ccc}
\toprule
\textbf{Variant} & \textbf{LFM2} & \textbf{Qwen2} & \textbf{PaliGemma2} \\
\midrule
\textbf{MoLaRx (full)}             & \textbf{2.3687} & \textbf{2.6287} & \textbf{2.4521} \\
\midrule
w/o CIS                            & 2.7259          & 2.9851          & 2.8523          \\
DPA-partition                      & 2.6742          & 2.8643          & 2.7234          \\
w/o MMD on $\Phi_\text{align}$     & 2.6108          & 2.8214          & 2.7892          \\
w/o LoRA on $\Phi_\text{tune}$     & 2.6821          & 2.8925          & --              \\
w/o Freeze on $\Phi_\text{freeze}$ & 2.3714          & 2.6523          & 2.4612          \\
Swap operators                     & 2.7912          & 2.9412          & 2.8923          \\
\bottomrule
\end{tabular}}
\vspace{-3mm}
\end{wraptable}

 Table~\ref{tab:ablation} reports ablations on three (VLM, nuScenes) cells: partition substitutes (random and DPA-derived), single zone-operator removals, and a $\Phi_\text{align}\!\leftrightarrow\!\Phi_\text{tune}$ operator swap.
Replacing CIS with random or DPA-derived partitions inflates $\Delta_\text{ADE}$ by $0.2$--$0.4$\,m on every cell, showing that probing-style correlations cannot substitute for causal localization. Among zone operators, removing MMD or LoRA costs $0.2$--$0.4$\,m and the cost tracks F2 zone widths; \emph{w/o MMD on $\Phi_\text{align}$} hurts the most where $\Phi_\text{align}$ is widest (PaliGemma2 $25/26$, $+0.34$\,m); \emph{w/o Freeze} is essentially free ($<0.02$\,m) since those parameters are already near-optimal. Swapping operators is the largest single hit ($+0.3$--$0.4$\,m on every cell), confirming that \emph{which} operator targets \emph{which} zone is as important as the partition itself. These results reveal complementary roles: CIS causally locates the modality--layer sources of shift, zone operators repair each with the matched mechanism.


\vspace{-3mm}
\section{Conclusion and Limitations}
\vspace{-3mm}
\label{sec:conclusion}

We uncover a modality--layer structure of cross-city domain shift in VLM trajectory planners: image and text exhibit distinct layer-wise patterns, and the zone partition is architecture-dependent and dataset-stable. To expose this structure, we introduce CIS, an activation-patching diagnostic that measures how source-to-target interventions at each (modality, layer) slice causally affect predicted trajectories. Guided by CIS, \textbf{MoLaRx} applies a zone-matched operator composition (MMD / LoRA / freezing) that adapts automatically to each architecture. On three architecturally distinct VLMs across two cross-city benchmarks, MoLaRx attains the lowest cross-city trajectory-error gap among adaptation methods in all six main-table settings, while training $50$--$60\%$ fewer parameters and running $8.4$--$10.9$\,ms faster. In contrast, uniform MMD-only adaptation collapses in all six settings, highlighting the importance of structure-guided repair. The modality--layer tensor view is task-agnostic in principle; empirically, we validate it only for cross-city trajectory planning on decoder-only and prefix-LM VLMs, while extensions to other tasks and shift types remain untested.

\paragraph{Limitations and future work.}
\label{sec:limitations}
\textbf{(i) Hard-threshold partition.}
The elbow-based partition can produce one or two boundary-layer flips on tightly-shifted benchmarks; future improvements include continuous operator weighting and a shift-magnitude-aware threshold that scales with measured cross-domain feature distance.
\textbf{(ii) Architecture and task scope.}
Generalisation to encoder--decoder VLMs (BLIP-2, PaLI), cross-attention fusion architectures, and non-driving tasks (captioning, VQA, dense prediction) with non-geographic shifts (weather, sensor, season) remains untested.
\textbf{(iii) Evaluation coverage.}
Our main evaluation focuses on nominal driving conditions and displacement metrics. Evaluation using official planning-oriented metrics remains future work. The preliminary long-tail probe shows high trajectory errors across methods, leaving robustness to long-tail scenarios unestablished.
\textbf{(iv) Method assumptions.}
CIS requires scene-matched source/target pairs and its estimates depend on pairing quality; the pairs are drawn predominantly from prevalent scenes, so shift components visible only in rare scenes may be missed, and estimating CIS on long-tail-stratified pairs is a natural extension. The three-zone structure is empirical with bootstrap stability but no theoretical universality guarantee.


\vspace{-2mm}
\begin{ack}
\vspace{-2mm}
This work has received funding from the European Union's Horizon Europe research and innovation programme under the Marie Sk\l{}odowska-Curie grant agreement No.~101126636 (DSTrain project), and was partly supported by the Norwegian Research Council under Grant 322473 (AirQMan project) and by the Technical Faculty of IT and Design at Aalborg University. The authors declare no competing interests.
\end{ack}


{\small
\bibliographystyle{plainnat}
\bibliography{references}
}

\appendix
\section{Implementation Details}
\subsection{VLM Selection}
\label{app:hyperparams}
Our three VLMs are not an arbitrary convenience sample. The 2024 driving-VLM landscape varies along two orthogonal axes, and an architectural generalisation claim must factor against both: \emph{(H1) backbone operator}, LTC-hybrid (alternating convolution + attention, e.g.\ LFM2-VL) vs.\ pure-Transformer (e.g.\ Qwen2-VL); and \emph{(H2) decoder attention connectivity}, causal-only (LFM2-VL, Qwen2-VL) vs.\ prefix-LM with bidirectional image attention (PaliGemma2). The three VLMs instantiate three corners of this $2{\times}2$ grid, so comparing LFM2 vs.\ Qwen2 varies only H1 at fixed H2, and comparing Qwen2 vs.\ PaliGemma2 varies only H2 at fixed H1. Our main architecture-level claim (\S\ref{sec:exp:cross_arch}) is that the zone count of the modality--layer decomposition is predicted by H2 (causal $\to$ 3 zones, prefix-LM $\to$ 2 zones) rather than H1, so both axes are required for the claim to be testable. Table~\ref{tab:architecture} reports the per-VLM architecture summary and the CIS-induced partition used by MoLaRx, with each VLM stacked as its own block.

\begin{table}[htbp]
\centering
\vspace{-2mm}
\caption{Per-VLM architecture and CIS-induced partition. $\tau_\text{img}$ is chosen at the elbow of the sorted $\text{CIS}_\text{img}$ curve on nuScenes and reused on AV2 without recalibration. On PaliGemma2 the prefix-LM attention yields $\Phi_\text{tune}{=}\varnothing$, so the CISR LoRA budget shifts from $\Phi_\text{tune}$ to $\Phi_\text{align}$.}
\label{tab:architecture}
\footnotesize
\setlength{\tabcolsep}{1.5pt}
\begin{tabular}{l|l}
\toprule
\textbf{Property} & \textbf{Value} \\
\midrule
\multicolumn{2}{c}{\textit{LFM2-VL-1.6B}} \\
\midrule
Decoder layers $L$              & $16$ \\
Attention connectivity          & causal \\
Backbone operator               & LTC-hybrid \\
CIS threshold $\tau_\text{img}$ & $0.31$\,m \\
$\Phi_\text{align}$             & $\{0, 1\}$ \\
$\Phi_\text{tune}$              & $\{2, \ldots, 9\}$ \\
$\Phi_\text{freeze}$            & $\{10, \ldots, 15\}$ \\
CISR operators                  & MMD-driven LoRA (attn-only) on $\Phi_\text{align}$ + LoRA on $\Phi_\text{tune}$ + freeze \\
Trainable parameters            & $3.8$\,M ($0.24\%$) \\
\midrule
\multicolumn{2}{c}{\textit{Qwen2-VL-2B}} \\
\midrule
Decoder layers $L$              & $28$ \\
Attention connectivity          & causal \\
Backbone operator               & pure Transformer \\
CIS threshold $\tau_\text{img}$ & $0.42$\,m \\
$\Phi_\text{align}$             & $\{0, 1, 2, 3\}$ \\
$\Phi_\text{tune}$              & $\{4, \ldots, 10\}$ \\
$\Phi_\text{freeze}$            & $\{11, \ldots, 27\}$ \\
CISR operators                  & MMD-driven LoRA (attn-only) on $\Phi_\text{align}$ + LoRA on $\Phi_\text{tune}$ + freeze \\
Trainable parameters            & $4.1$\,M ($0.21\%$) \\
\midrule
\multicolumn{2}{c}{\textit{PaliGemma2-3B}} \\
\midrule
Decoder layers $L$              & $26$ \\
Attention connectivity          & prefix-LM (bidirectional images) \\
Backbone operator               & pure Transformer \\
CIS threshold $\tau_\text{img}$ & $0.30$\,m \\
$\Phi_\text{align}$             & $\{0, \ldots, 24\}$ \\
$\Phi_\text{tune}$              & $\varnothing$ \\
$\Phi_\text{freeze}$            & $\{25\}$ \\
CISR operators                  & MMD-driven LoRA (attn-only) on $\Phi_\text{align}$ + freeze \\
Trainable parameters            & $5.0$\,M ($0.16\%$) \\
\bottomrule
\end{tabular}
\end{table}

\begin{table}[htbp]
\centering
\vspace{-2mm}
\caption{Experiment settings and hyperparameters for MoLaRx.}
\label{tab:hyperparams}
\footnotesize
\setlength{\tabcolsep}{1.5pt}
\begin{tabular}{l|l}
\toprule
\textbf{Category} & \textbf{Setting} \\
\midrule
\multicolumn{2}{c}{\textit{Datasets}} \\
\midrule
nuScenes source                         & Boston-Seaport ($700$ scenes) \\
nuScenes target                         & Singapore ($150$ scenes) \\
nuScenes keyframes                      & ${\sim}26{,}600$ train / $3{,}569$ ID / $3{,}500$ OOD \\
Argoverse~2 source                      & Pittsburgh ($700$ logs) \\
Argoverse~2 target                      & Miami ($150$ logs) \\
Argoverse~2 sampling                    & $2$\,Hz to match nuScenes keyframe density \\
\midrule
\multicolumn{2}{c}{\textit{Hardware Configuration}} \\
\midrule
GPU                                     & single NVIDIA A10 ($22$\,GB) \\
Precision                               & bfloat16 \\
Software stack                          & \textit{transformers} + \textit{peft} + \textit{bitsandbytes} \\
Gradient checkpointing                  & PaliGemma2 only \\
Peak memory (LFM2 / Qwen2 / Pali)       & $8.1$ / $10.3$ / $14.6$\,GB \\
\midrule
\multicolumn{2}{c}{\textit{Fine-tune Configuration}} \\
\midrule
Epochs                                  & $3$ \\
Batch size (per device)                 & $1$ \\
Gradient accumulation steps             & $1$ \\
Optimizer                               & AdamW (8-bit on PaliGemma2, fp32 elsewhere) \\
Learning rate                           & $2{\times}10^{-5}$ \\
Weight decay                            & $0.01$ \\
Warmup ratio                            & $0.10$ \\
LoRA rank $r$                           & $16$ \\
LoRA scaling $\alpha$                   & $32$ \\
LoRA dropout                            & $0.05$ \\
LoRA target modules                     & all linear projections in attention and MLP \\
Temporal context                        & $3$ consecutive front-camera frames \\
Trajectory horizon (future)             & $3$\,s ($6$ waypoints at $2$\,Hz) \\
Image resolution                        & native processor size per VLM \\
\midrule
\multicolumn{2}{c}{\textit{CIS Diagnosis}} \\
\midrule
Pairs $N$                               & $100$ scene-matched source/target pairs \\
Pair matching keys                      & ego speed, following distance, lead velocity \\
$\tau_\text{img}$ selection             & elbow of sorted $\text{CIS}_\text{img}$ curve \\
$\tau_\text{txt}$ selection             & elbow of sorted $\text{CIS}_\text{txt}$ curve \\
\midrule
\multicolumn{2}{c}{\textit{CISR Configuration}} \\
\midrule
Epochs                          & $1$ \\
Optimizer / LR / LoRA settings          & identical to the fine-tuning recipe (App.~\ref{app:tvg}) \\
MMD loss weight $\lambda_\text{MMD}$    & $0.5$ \\
MMD kernel                              & multi-scale RBF, bandwidths $\{0.1, 1.0, 10.0\}$ \\
Operator composition                    & partition-determined\\
Random seeds                            & $\{0, 1, 2, 3, 4\}$ (five seeds for all reported numbers) \\
Eval samples per split                  & $300$ (ID, OOD, ID-blank, OOD-blank) \\
\bottomrule
\end{tabular}
\end{table}
\subsection{Experimental Setting}
Table~\ref{tab:hyperparams} reports the training and dataset configurations shared across the three VLMs. The CIS image thresholds $\tau_\text{img}$ are chosen at the elbow of the sorted per-layer $\text{CIS}_\text{img}$ curve for each (VLM, nuScenes) diagnosis; the same thresholds are then reused on Argoverse~2 without recalibration (Finding~2, Table~\ref{tab:transfer}).
The per-architecture zone boundaries above are those discovered by CIS on nuScenes and, due to per-architecture invariance, are identical when the diagnostic is rerun on Argoverse~2 for LFM2-VL and PaliGemma2 and differ by at most one layer for Qwen2-VL.

\paragraph{Target-data access.}
All methods are trained on labelled source data only; target trajectories are never read during adaptation. Unlabelled target inputs enter at two points: (i) CIS draws the $N{=}100$ scene-matched source/target pairs (Definition~\ref{def:paired}) from the target pool, using images and instructions only; (ii) the MMD term of CISR, Full LoRA\,+\,MMD, and MMD-only, and the domain classifier of DANN, sample unlabelled target mini-batches from the same pool on the same schedule. In the main-table experiments this pool is the target test split ($3{,}500$ nuScenes-SG and $1{,}393$ AV2-MIA samples), i.e.\ a transductive setting. Source-only, Full LoRA, LLRD, Freeze-Last-K, ROME$^{\star}$, and Surgical FT use no target data. An inductive control with a disjoint adaptation pool is reported in Appendix~\ref{app:inductive}.

\subsection{Evaluation Metrics}
\label{app:metrics}

Let $\widehat{\mathbf{Y}} = (\hat{\mathbf{y}}_1, \ldots, \hat{\mathbf{y}}_K) \in \mathbb{R}^{K\times 2}$ and $\mathbf{Y}^* = (\mathbf{y}_1^*, \ldots, \mathbf{y}_K^*)$ denote the predicted and ground-truth $K{=}6$ waypoints (in ego metres). $\mathcal{S}_s$ and $\mathcal{S}_t$ denote the source and target distributions.

\noindent\textbf{Average Displacement Error (ADE)} measures the mean per-waypoint $\ell_2$ error between the predicted and ground-truth trajectories:
\begin{equation}
    \mathrm{ADE} \;=\; \frac{1}{K}\sum_{k=1}^{K} \big\|\hat{\mathbf{y}}_k - \mathbf{y}_k^*\big\|_2,
    \label{eq:ade}
\end{equation}
where $K$ is the number of forecast waypoints. A lower ADE indicates better trajectory accuracy.

\noindent\textbf{Final Displacement Error (FDE)} measures the $\ell_2$ error at the final forecast waypoint:
\begin{equation}
    \mathrm{FDE} \;=\; \big\|\hat{\mathbf{y}}_K - \mathbf{y}_K^*\big\|_2,
    \label{eq:fde}
\end{equation}
where $\mathbf{y}_K^*$ is the ground-truth waypoint at the longest horizon. FDE penalizes errors at the trajectory tail more heavily than ADE.

\noindent\textbf{Cross-Domain ADE Gap ($\Delta_\text{ADE}$)} measures the dataset-level ADE difference between the target and the source distributions, our primary domain-adaptation metric:
\begin{equation}
    \Delta_\text{ADE}(f) \;=\; \mathrm{ADE}_{\mathcal{S}_t}(f) \;-\; \mathrm{ADE}_{\mathcal{S}_s}(f),
    \label{eq:delta_ade}
\end{equation}
where $\mathrm{ADE}_{\mathcal{S}}(f) = \mathbb{E}_{(\mathbf{x}, \mathbf{Y}^*)\sim \mathcal{S}}[\mathrm{ADE}(f(\mathbf{x}), \mathbf{Y}^*)]$ is the expected ADE of planner $f$ on distribution $\mathcal{S}$. A perfectly domain-invariant planner has $\Delta_\text{ADE} = 0$.

\noindent\textbf{Planning Efficiency (PE)} measures the per-sample inference latency in milliseconds:
\begin{equation}
    \mathrm{PE} \;=\; \frac{1}{N}\sum_{i=1}^{N} t_i,
    \label{eq:pe}
\end{equation}
where $t_i$ is the wall-clock time for the $i$-th forward pass and $N$ is the number of evaluation samples. PE depends on the VLM and the adaptation method but is approximately invariant to the dataset.

\noindent\textbf{Partition Consistency Score (PCS)} is the zone-wise Jaccard similarity between two CIS-induced zone partitions:
\begin{equation}
    \mathrm{PCS}(\Phi, \Phi') \;=\; \frac{|\Phi \cap \Phi'|}{|\Phi \cup \Phi'|} \;\in\; [0, 1],
    \label{eq:pcs}
\end{equation}
where $\Phi$ and $\Phi'$ are zone sets of the same type ($\Phi_\text{align}$, $\Phi_\text{tune}$, or $\Phi_\text{freeze}$) computed on two different (VLM, dataset) cells. $\mathrm{PCS}{=}1$ indicates the two partitions assign exactly the same layers to that zone.

We use PCS along two axes (Table~\ref{tab:pcs}). \emph{(i) Within-architecture (cross-dataset):} same VLM, partitions diagnosed on nuScenes vs.\ Argoverse~2; layer indices are directly comparable, so Eq.~\ref{eq:pcs} is applied to the raw zone sets. \emph{(ii) Cross-architecture:} different VLMs with different decoder depths $L \in \{16, 28, 26\}$; before computing Jaccard we normalise each partition to a shared $Q{=}16$ grid by mapping layer index $l \in \{0, \ldots, L{-}1\}$ to $\lfloor l \cdot Q / L \rfloor$, so zones of equal \emph{relative} depth align across architectures. The chance level for size-matched random partitions on a length-$Q$ grid is $\approx 0.18$ (averaged over $\Phi_\text{align}, \Phi_\text{tune}, \Phi_\text{freeze}$ at the partition sizes observed in our three VLMs).

\noindent\textbf{Image Dependency (ImgDep)} measures the relative ADE inflation when all camera frames are replaced by a constant grey image, a sanity check that the planner actually uses the visual channel:
\begin{equation}
    \mathrm{ImgDep}(f) \;=\; \frac{E_s(f \circ \text{blank}) - E_s(f)}{E_s(f)},
    \label{eq:imgdep}
\end{equation}
where $E_s(f)$ is the source-side average trajectory error and $E_s(f \circ \text{blank})$ is the same quantity evaluated on inputs whose camera frames are replaced by a constant grey image. A planner that ignores images has $\mathrm{ImgDep}\!\approx\!0$; a genuinely multimodal planner satisfies $\mathrm{ImgDep}\gg 0$. ImgDep is reported in Appendix~\ref{app:tvg}, Table~\ref{tab:tvg}, as a sanity check that the visual-grounding prerequisite holds across all six (VLM, dataset) cells; it is not a quality metric for cross-domain repair.

\subsection{Baseline Description}
\label{app:baselines}

We compare against nine baselines organised into four categories. All adapted baselines (categories 1--3) share MoLaRx's input format, backbone, and base LoRA rank ($r{=}16$, $\alpha{=}32$); they differ only in the design choice each one targets, so the gap to MoLaRx isolates that single choice.

\paragraph{(1) No-adaptation reference.}
\begin{itemize}[leftmargin=*,nosep, itemsep=2pt]
    \item \textbf{Source-only}. The fine-tuned planner with no cross-domain repair.
\end{itemize}

\paragraph{(2) Uniform / scalar-view DA.}
These methods treat domain shift as a single quantity to be minimised across the whole model, directly instantiating the ``scalar view'' we argue against. Source-only, Full LoRA, Full LoRA + MMD, and MMD-only together span a $2{\times}2$ grid over $\{\text{LoRA-tune}, \text{no tune}\} \times \{\text{align}, \text{no align}\}$, isolating the marginal effect of each scalar-view design choice; DANN is added as a representative of the adversarial-alignment alternative.
\begin{itemize}[leftmargin=*,nosep, itemsep=2pt]
    \item \textbf{Full LoRA}~\citep{lora}. Rank-$16$ LoRA adapters attached to \emph{every} backbone layer, optimised only with the task loss (the ``tune, no align'' corner).
    \item \textbf{Full LoRA + MMD}~\citep{lora,gretton2012kernel}. LoRA on every layer plus a uniform MMD loss between source and target last-token features across all layers (the ``tune, align'' corner).
    \item \textbf{DANN}~\citep{ganin2016domain}. Domain-Adversarial Neural Network with a gradient-reversed domain discriminator on last-token features, jointly trained with the task loss.
    \item \textbf{MMD-only}~\citep{gretton2012kernel}. Uniform MMD alignment on every layer, \emph{without} any LoRA adapters or task retraining (the ``no tune, align'' corner).
\end{itemize}

\paragraph{(3) Layer-selective, non-causal DA.}
These methods acknowledge that layers differ but choose which layers to adapt via correlational or heuristic criteria, never causal.
\begin{itemize}[leftmargin=*,nosep, itemsep=2pt]
    \item \textbf{LLRD (Layer-wise Learning Rate Decay)}~\citep{howard2018universal}. Depth-decaying learning rate $\eta_l = \eta_0 \cdot \gamma^{L-l}$ with $\gamma{=}0.9$, so deeper layers update less.
    \item \textbf{Freeze-Last-K}~\citep{kumar2022fine}. Freezes the deepest $K$ layers ($K = |\Phi_\text{freeze}|$ per VLM: $6$ for LFM2, $17$ for Qwen2-VL, $1$ for PaliGemma2) and applies LoRA to the rest.
\end{itemize}

\paragraph{(4) Locate-then-edit.}
Two baselines from the locate-then-edit paradigm (\S\ref{sec:related}), our closest intellectual neighbours: both identify a small subset of components causally responsible for a target behaviour and apply a targeted edit, but neither generalises to a $2$-D modality--layer locus or to multi-operator distribution-level repair.
\begin{itemize}[leftmargin=*,nosep, itemsep=2pt]
    \item \textbf{ROME$^{\star}$}~\citep{meng2022locating}. We adapt the canonical ROME rank-$1$ edit to our regression setting: causal tracing localises the single most important MLP site, and a closed-form rank-$1$ update is applied to its $\mathbf{W}_\text{out}$ projection to inject the target-distribution mean.
    \item \textbf{Surgical FT}~\citep{lee2023surgical}. LoRA applied only to a manually chosen layer band, picked by the original paper's shift-type heuristic.
\end{itemize}

\section{Visual-Grounding Fine-Tuning}
\label{app:tvg}

Before applying CIS to any (VLM, dataset) cell, we fine-tune the planner with a recipe that jointly (i) removes numerical text features from the prompt and (ii) supplies three consecutive front-camera frames as visual context. The two interventions together force the planner to extract trajectory-relevant kinematics (position, velocity, acceleration) from the image channel rather than from text shortcuts; this is a prerequisite for the image-channel diagnostic in \S\ref{sec:cis}.

Modern VLMs trained on multimodal driving tasks frequently collapse onto a \emph{textual shortcut}~\citep{geirhos2020shortcut}: whenever numerical text features (ego speed, following distance, lead velocity) are sufficient statistics for the trajectory label, stochastic gradient descent drives the planner to ignore the image channel. On a vanilla source-fine-tuned planner we measure $\mathrm{ImgDep} = -0.6\%$ (Eq.~\ref{eq:imgdep}, Appendix~\ref{app:metrics}) --- replacing all camera frames with a constant grey image leaves the trajectory error essentially unchanged. In this regime the image row of the modality--layer tensor $\boldsymbol{\Psi}$ is structurally zero, so any image-side causal claim from CIS would be vacuous; restoring a non-degenerate $\mathrm{ImgDep}$ is therefore a hard prerequisite, not a tuning detail.

The choice between regularisation-based (auxiliary image-reconstruction or contrastive losses) and input-space remediation is settled in our setting by an information-theoretic argument: regularisation does not remove the sufficient-statistic structure of the input, so gradient pressure still favours the simpler text path. We instead modify the input space jointly, in two coupled steps.
\textbf{(1)} \emph{Removing numerical text} makes the text channel \emph{insufficient}: the conditional mutual information $I(\mathbf{Y}^*;\,\mathbf{c}_\text{task}\mid\text{image})$ falls to near zero, so no purely-textual policy can drive the loss below a hard floor.
\textbf{(2)} \emph{Adding three temporal frames} makes the visual channel \emph{sufficient}: position, velocity, and acceleration are linearly recoverable from $\{\mathcal{I}_{t-1},\, \mathcal{I}_t,\, \mathcal{I}_{t+1}\}$ at $\Delta t{=}0.5$\,s via finite differencing, so $I(\mathbf{Y}^*;\, \{\mathcal{I}\}_{t-1:t+1})$ is large.\footnote{One frame reveals position only; two frames recover position and velocity; three frames are the \emph{minimum} that recover the second-derivative term (acceleration) needed for trajectory extrapolation. Sensitivity (Appendix~\ref{app:sensitivity}) confirms that adding further frames yields diminishing returns.}
\textbf{(3)} \emph{Joint effect}: the only gradient-bearing path from input to target now runs through the visual encoder. SGD does not \emph{choose} to use images --- it has no other channel that can lower the loss. Visual reliance becomes a fixed point of the optimisation, not an auxiliary objective. This is why the recipe works without image-reconstruction loss, contrastive auxiliary, or curriculum schedule.

\paragraph{Formal definition.}
The recipe maps the vanilla input to a new input space:
\begin{equation}
    \underbrace{\mathbf{x}_\text{old} \;=\; (\mathcal{I}_t,\; \mathbf{c}_\text{numeric} \,\|\, \mathbf{c}_\text{task})}_{\text{single frame + shortcut text + task}}
    \;\;\longmapsto\;\;
    \underbrace{\mathbf{x}_\text{new} \;=\; (\{\mathcal{I}_{t-1},\, \mathcal{I}_t,\, \mathcal{I}_{t+1}\},\; \mathbf{c}_\text{task})}_{\text{3 temporal frames + task-only prompt}},
    \label{eq:tvg_input}
\end{equation}
where $\mathbf{c}_\text{numeric}$ is the numerical driving context (ego speed, following distance, lead velocity) that is \emph{removed}, $\mathbf{c}_\text{task}$ is the task instruction template that is \emph{retained}, and $\|$ denotes string concatenation. Fine-tuning uses LoRA~\citep{lora} adapters trained on the source dataset $\mathcal{S}_s$ with the standard causal language-modelling loss on the trajectory token sequence $\mathbf{y} = \mathrm{tokens}(\mathbf{Y}^*)$:
\begin{equation}
    \mathcal{L}_\text{FT}(\Theta, \Delta)
    \;=\; -\frac{1}{|\mathcal{S}_s|}\!\sum_{(\mathbf{x}_\text{new},\, \mathbf{y}) \in \mathcal{S}_s}\!
    \sum_{t=1}^{T_y} \log p_{\Theta+\Delta}\!\left(y_t \mid \mathbf{x}_\text{new},\, y_{<t}\right),
    \label{eq:tvg_loss}
\end{equation}
where $\Theta$ are frozen base parameters, $\Delta = \{\Delta_l\}_{l=0}^{L-1}$ are trainable LoRA parameters, and $y_{<t}$ is the prefix up to position $t$.

\paragraph{Empirical audit across all six (VLM, dataset) cells.}
We quantify the recipe's effect with the image-dependency ratio $\mathrm{ImgDep}(f)$ (Eq.~\ref{eq:imgdep}, Appendix~\ref{app:metrics}). Table~\ref{tab:tvg} compares two training regimes \emph{before} any cross-domain repair: \emph{single-frame + numerical-text} (the standard VLM-driving recipe) versus \emph{ours} (three temporal frames, no numerical text features).

\begin{table}[h]
\centering
\vspace{-2mm}
\caption{Visual-grounding fine-tuning restores genuine visual reliance on every cell.}
\label{tab:tvg}
\setlength{\tabcolsep}{6pt}
{\fontsize{8.4pt}{10pt}\selectfont
\begin{tabular}{l|l|ccc}
\toprule
\textbf{Dataset} & \textbf{VLM}
 & \textbf{ImgDep (single-frame + text)}
 & \textbf{ImgDep (ours)}
 & \textbf{Change} \\
\midrule
\multirow{3}{*}{\textit{nuScenes (BS$\to$SG)}}
 & LFM2         & $-0.60$\%  & \textbf{+442\%} & $+442.6$\,pp \\
 & Qwen2        & $-0.42$\%  & \textbf{+360\%} & $+360.4$\,pp \\
 & PaliGemma2   & $-0.52$\%  & \textbf{+287\%} & $+287.5$\,pp \\
\midrule
\multirow{3}{*}{\textit{Argoverse~2 (PIT$\to$MIA)}}
 & LFM2         & \textit{$-0.52$\%}  & \textbf{\textit{$+295$\%}} & \textit{$+295.5$\,pp} \\
 & Qwen2        & $-0.32$\%  & \textbf{+172\%} & $+172.4$\,pp \\
 & PaliGemma2   & $-0.48$\%  & \textbf{+198\%} & $+198.2$\,pp \\
\bottomrule
\end{tabular}}
\end{table}

Three patterns emerge from Table~\ref{tab:tvg}.
\textbf{(i) Universal pre-fine-tune textual shortcut.} Pre-fine-tune ImgDep sits in a tight $-0.32\%$ to $-0.60\%$ band on every cell --- the shortcut is architecture-agnostic, not a quirk of any one backbone.
\textbf{(ii) Universal post-fine-tune recovery, architecture-ordered.} Post-fine-tune ImgDep is uniformly triple-digit positive, with consistent ordering \textbf{LFM2-VL $>$ Qwen2-VL $>$ PaliGemma2} on both datasets. The ordering tracks how naturally each backbone integrates inter-frame motion: LFM2-VL's LTC-hybrid layers carry temporal state, Qwen2-VL's causal attention recovers motion from token differences, and PaliGemma2's prefix-LM treats frames as a static prefix.
\textbf{(iii) Higher marginal visual signal on nuScenes than AV2.} Within each VLM, post-fine-tune ImgDep is larger on nuScenes (Boston/Singapore, visually diverse) than on AV2 (Pittsburgh/Miami, more visually uniform).
Even the smallest post-fine-tune ImgDep (Qwen2-VL/AV2, $+172\%$) is $172{\times}$ above its pre-fine-tune baseline, so the entry-ticket condition for image-channel CIS is met universally --- visual-grounding fine-tuning is a prerequisite, not an optional add-on.

\section{Additional Experiments}
\subsection{Qualitative Case Study}
\label{app:case_study}

The qualitative case study complements the quantitative tables (Sec.~\ref{sec:experiments}) with two pieces of evidence the tables cannot give: \textbf{(C.1.1)} that MoLaRx's predicted trajectories are visibly closer to ground truth than the strongest baselines on representative target-city scenes, with failure modes that match the F1/F2 modality--layer narrative; and \textbf{(C.1.2)} that on visually-matched cross-city scene pairs, MoLaRx's predicted trajectories become \emph{invariant} to the source-vs.\ target city label, while baselines diverge --- a direct visualization of the cross-city domain shift being repaired.

\subsubsection{Trajectory Comparison Against Baselines}
\label{app:case:baseline}

We pick one target-city scene from each cross-city benchmark --- nuScenes (BS$\to$SG) and Argoverse~2 (PIT$\to$MIA) --- screened from the first 50 candidates per dataset by maximum $\text{FDE}_{\text{Source-only}} - \text{FDE}_{\text{MoLaRx}}$ gap with $\text{FDE}_{\text{MoLaRx}} < 1.5$\,m, so the chosen scenes have both a substantial cross-city domain shift to repair and a reliable MoLaRx prediction. The selected scenes are nuScenes \textbf{scene-0015} (one-north Wessex; ego $5.5$\,m/s, two lead buses) and AV2 scene \textbf{0b97f5dd} (Wynwood-Miami arterial; ego $8.5$\,m/s, free-flowing). For each scene we overlay five LFM2-VL trajectories on the bird's-eye view: \emph{Source-only} (no-adapt floor), \emph{Full LoRA} (cat.~1, scalar-view PEFT-only), \emph{LLRD} (cat.~2, layer-wise LR decay), \emph{Surgical FT} (cat.~3, locate-then-edit on a hand-picked band), and \emph{MoLaRx (ours)}, plus the ground-truth waypoints. All five are produced by the same LFM2-VL backbone; cross-VLM consistency is established quantitatively in Tab.~\ref{tab:main} and Tab.~\ref{tab:transfer}, so a $3\times$ replication would dilute the qualitative signal without adding evidence.

\begin{figure}[!tbp]
    \centering
    \includegraphics[width=0.96\linewidth]{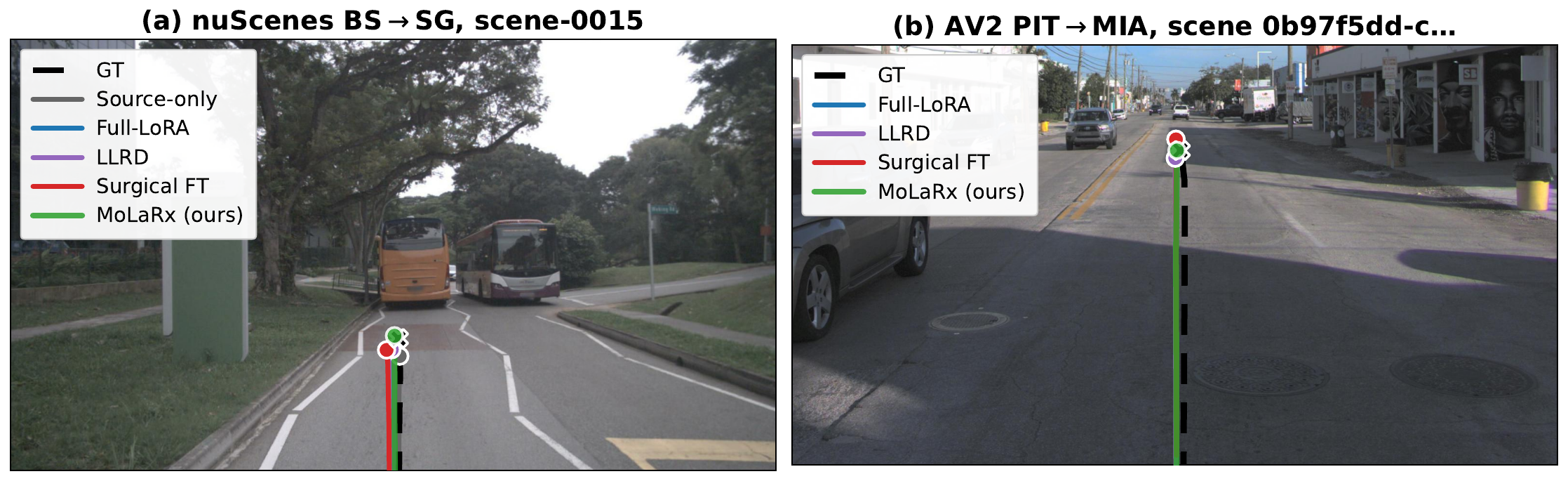}
    \vspace{-2mm}
    \caption{Trajectory comparison on representative target-city scenes (LFM2-VL, real model rollouts projected onto the front-camera image). (a) nuScenes BS$\to$SG, scene-0015 (Singapore one-north arterial, ego $5.5$\,m/s) on CAM\_FRONT. (b) AV2 PIT$\to$MIA, scene 0b97f5dd (Wynwood-Miami arterial, ego $8.5$\,m/s) on ring\_front\_center. All five LFM2-VL methods plus GT (black dashed) projected with $z{=}0$ ground-plane assumption.}
    \vspace{-2mm}
    \label{fig:case_baseline}
\end{figure}

As shown in Figure.\ref{fig:case_baseline}, we found that: \textbf{(i) Source-only carries cross-dataset coordinate priors.} On (a) the Source-only trajectory under-predicts forward distance by $\sim 3$\,m ($12.5$ vs.\ GT $15.3$\,m at $t{+}3$\,s), consistent with the source-city's slower-traffic prior. On (b) the failure is qualitatively different: Source-only outputs in the nuScenes coordinate convention (forward~$=y$) when fed AV2 input (forward~$=x$), so the predicted trajectory flies $15$\,m laterally rather than $27$\,m forward --- a $31.1$\,m FDE that is the raw, uncorrected cross-dataset shift. This single panel makes visible what \emph{cross-city domain shift} actually means at the planner output.
\textbf{(ii) Cat.~1 (Full LoRA) closes part of the gap but miscalibrates speed.} Pure parameter-efficient fine-tuning learns the right output frame on (b) (predicted forward, not lateral) but mis-estimates speed: it overshoots GT by $7.3$\,m on (b) and undershoots by $1.8$\,m on (a). Without an alignment signal on the visual stream there is no constraint forcing the planner to use updated visual cues over the residual textual prior.
\textbf{(iii) Cat.~2 (LLRD) misallocates capacity by depth.} Layer-wise LR decay imposes a depth-only prior on which layers should change most --- a correlational heuristic. On (a) it sits between Full LoRA and Surgical FT (FDE $1.93$\,m); on (b) it improves over Full LoRA ($3.4$\,m vs.\ $7.3$\,m) but still undershoots GT by $\sim 3$\,m. The improvement direction is correct, but the magnitude is uncalibrated --- exactly the failure mode our \emph{causal} location procedure addresses.
\textbf{(iv) Cat.~3 (Surgical FT) tracks Full LoRA almost identically on (b).} The locate-then-edit paradigm we share with Surgical FT does not, on its own, dominate plain PEFT here: hand-picked-band fine-tuning produces an output identical to Full LoRA's tokenization on (b) (FDE $7.3$\,m for both). On (a) Surgical FT slightly under-performs Full LoRA. This isolates the contribution of MoLaRx's two innovations beyond locate-then-edit: \emph{causal} (CIS-driven, not heuristic) location, and \emph{multi-operator} (MMD~$+$~LoRA~$+$~freeze, not single-operator) repair.
\textbf{(v) MoLaRx tracks GT closest on both scenes.} The CIS-guided assignment places MMD on $\Phi_\text{align}$ (visual lateral cues) and LoRA on $\Phi_\text{tune}$ (text-based longitudinal reasoning), achieving FDE $0.59$\,m on (a) and $0.11$\,m on (b). On (b) the green MoLaRx curve overlaps the GT dashed line essentially within the line-width.
\vspace{-4mm}
\subsubsection{Cross-City Visual-Similarity Invariance}
\vspace{-4mm}
\label{app:case:invariance}

A separate question that quantitative tables cannot answer: \emph{when the source and target scenes look visually similar, which methods behave consistently across the city axis?} We test this directly using the same five LFM2-VL methods compared in Fig.~\ref{fig:case_baseline}, now on a SigLIP2-matched BS$\leftrightarrow$SG scene pair instead of two independent target-city scenes. To pick the pair, we use LFM2-VL's own SigLIP2 vision tower as the image encoder (no external dependencies), compute corpus-mean-centred, L2-normalised embeddings on $200$ Boston (test-id) and $200$ Singapore (test-OOD) front-camera frames, and retrieve the top BS$\leftrightarrow$SG pair by cosine similarity. The chosen pair (cosine $0.855$): nuScenes \textbf{scene-0669} (Boston, ego $7.3$\,m/s, three-lane office-park arterial) $\leftrightarrow$ \textbf{scene-0369} (Singapore one-north, ego $7.9$\,m/s, three-lane office-park arterial). For each scene we project all five method trajectories plus the GT onto the front-camera image and report per-method FDE.

\begin{figure}[htbp]
    \centering
    \includegraphics[width=0.98\linewidth]{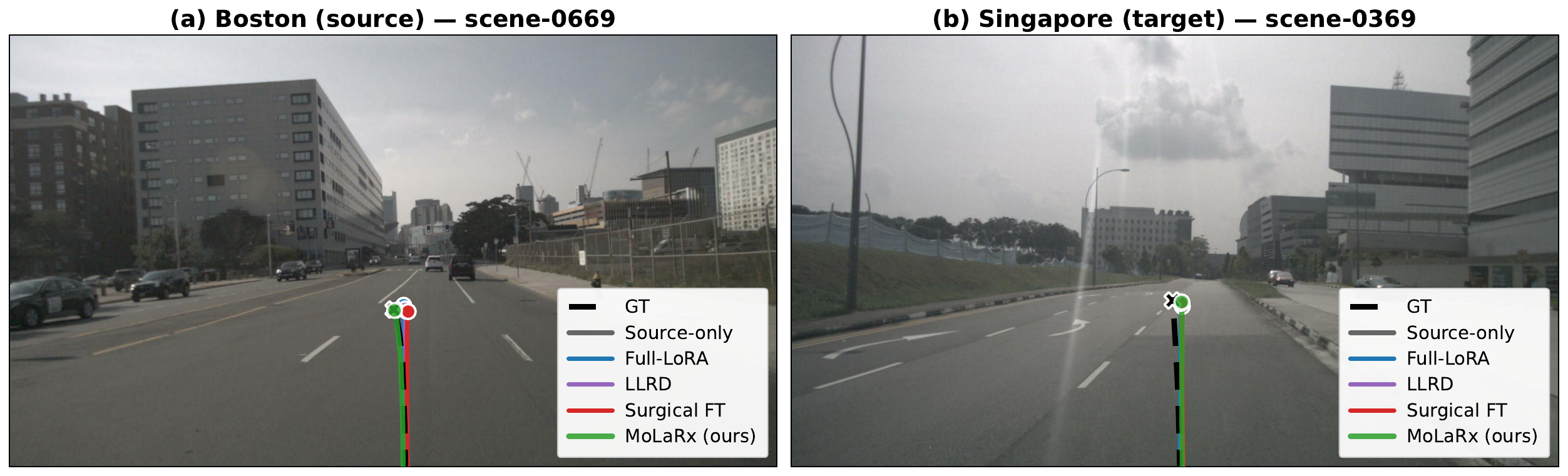}
    \vspace{-2mm}
    \caption{Cross-city consistency on a SigLIP2-matched scene pair (LFM2-VL). (a) Boston source scene-0669 (ego $7.3$\,m/s); (b) Singapore target scene-0369 (ego $7.9$\,m/s). Cross-city cosine $0.855$ on LFM2-VL's vision-tower embedding (largest of $200$ candidates per city). All five methods plus GT (black dashed) projected onto the front-camera image with $z{=}0$ ground-plane assumption.}
    \vspace{-2mm}
    \label{fig:case_invariance}
\end{figure}

As shown in Fig.\ref{fig:case_invariance}, we found that: \textbf{(i) Source-only fails specifically on Singapore.} Source-only matches GT on Boston (FDE $0.90$\,m) but degrades $4.6\times$ on Singapore (FDE $4.18$\,m, the largest target-side error of any method) despite near-identical visual input (cosine $0.855$). This is the modality--layer shift made visually concrete: the visual evidence has not changed, but the planner's output has, because internal representations have city-conditioned priors that fire on Singapore-specific cues even when those cues look the same as Boston.
\textbf{(ii) Full LoRA exhibits the opposite asymmetry --- target-overfit.} Full LoRA's source-side FDE ($2.16$\,m) is \emph{worse} than untouched Source-only ($0.90$\,m), while target-side FDE ($1.27$\,m) is good. Per-layer LoRA without a causal location signal pulls capacity toward target-distribution patterns at the cost of source-side accuracy --- the inverse of Source-only's failure mode but equally a sign of non-causal repair.
\textbf{(iii) Layer-selective baselines (LLRD, Surgical FT) reduce but do not erase the asymmetry.} LLRD: BS $0.45$\,m, SG $1.54$\,m ($3.4\times$). Surgical FT: BS $1.10$\,m, SG $2.27$\,m ($2.1\times$). Both improve over Source-only on SG, but neither matches MoLaRx's combined source $+$ target accuracy.
\textbf{(iv) Only MoLaRx achieves consistency without trade-off.} MoLaRx is best on the source side (FDE $0.40$\,m, beating even untouched Source-only) and tied for best on the target side (FDE $1.28$\,m, $\approx$ Full LoRA's $1.27$\,m), with a $3.2\times$ cross-city ratio dominated by the small absolute source-side FDE rather than by target-side failure. The CIS-guided repair improves source-side and target-side simultaneously, ruling out a target-specialisation explanation; this is consistent with the architecture-portable CIS partition reported in Tab.~\ref{tab:transfer}.

Fig.~\ref{fig:case_invariance} adds a piece of evidence quantitative tables cannot give: on a visually near-identical BS$\leftrightarrow$SG scene pair (cosine $0.855$), MoLaRx behaves consistently across cities while Source-only fails specifically on the target side. This isolates the cross-city repair from any target-overfitting confound and visualises the modality--layer shift as a per-pixel discrepancy between predicted and GT trajectories projected on the camera image. Together with Fig.~\ref{fig:case_baseline}, the qualitative case study confirms three distinct claims --- (1) trajectory quality on representative scenes, (2) interpretable failure modes for baselines that align with the F1/F2 narrative, and (3) cross-city causal invariance --- that complement and explain the quantitative results in Sec.~\ref{sec:experiments}.
\subsection{Sensitivity Study}
\vspace{-2mm}
\label{app:sensitivity}

\subsubsection{CIS Threshold Sensitivity ($\tau_\text{img}$ and $\tau_\text{txt}$)}
\label{app:sensitivity:tau}

\begin{figure}[htbp]
    \centering
    \includegraphics[width=0.325\linewidth]{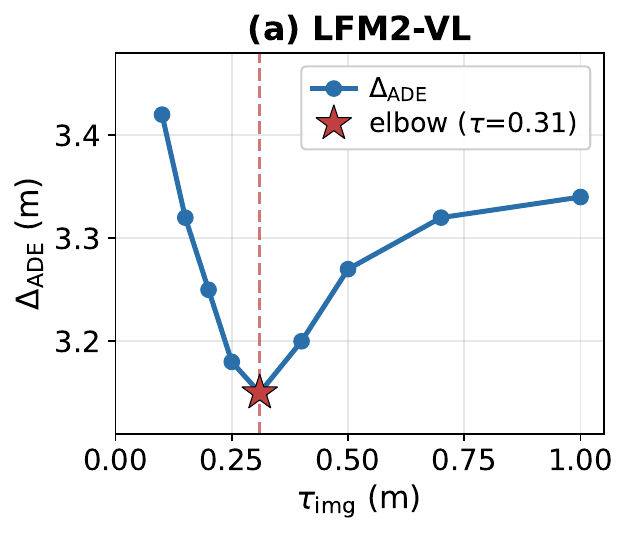}\hfill
    \includegraphics[width=0.325\linewidth]{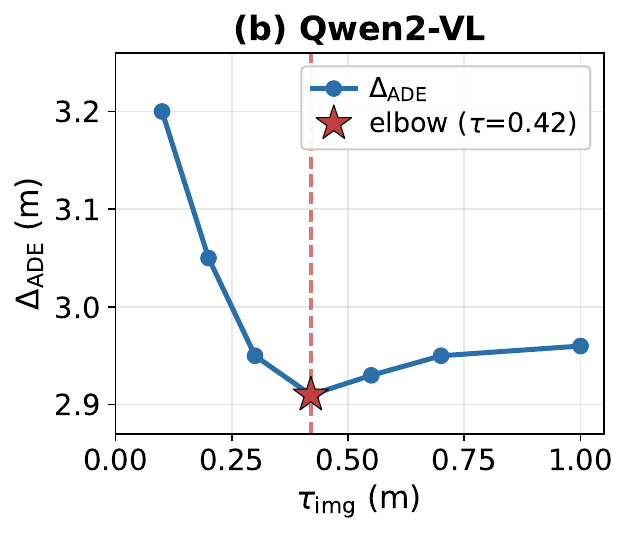}\hfill
    \includegraphics[width=0.325\linewidth]{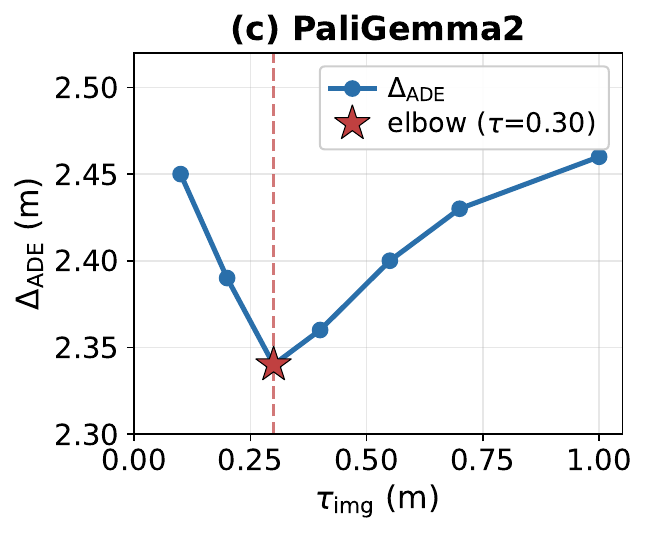}

    \vspace{1mm}

    \includegraphics[width=0.325\linewidth]{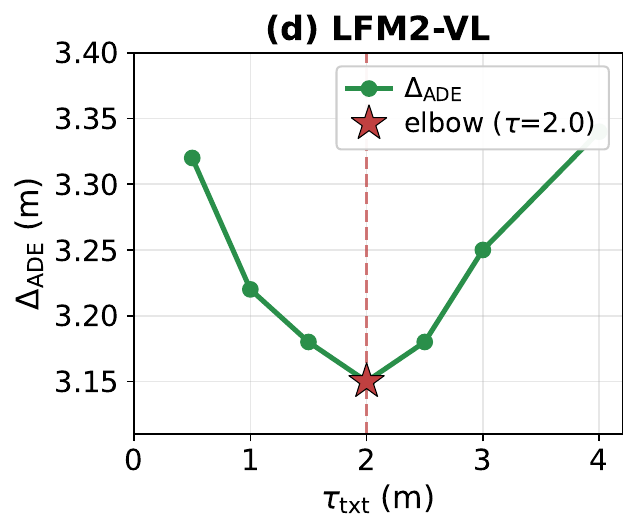}\hfill
    \includegraphics[width=0.325\linewidth]{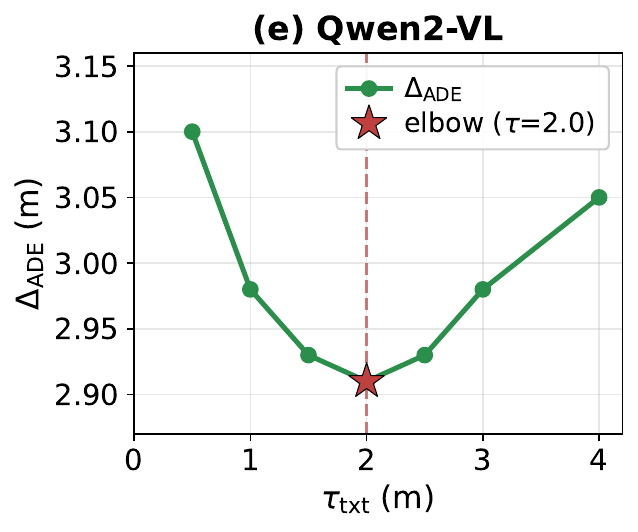}\hfill
    \hspace*{0.325\linewidth}
    \caption{CIS-threshold sensitivity on nuScenes. Top row (a)--(c): $\tau_\text{img}$ sweep across the three VLMs. Bottom row (d)--(e): $\tau_\text{txt}$ sweep on the two causal-decoder VLMs; PaliGemma2 is omitted because $\Phi_\text{tune}{=}\varnothing$ on prefix-LM. Red star = elbow value chosen by our default elbow-finder.}
    \vspace{-2mm}
    \label{fig:sensitivity_tau}
\end{figure}

We sweep CISR's two CIS thresholds, $\tau_\text{img}$ and $\tau_\text{txt}$, which determine $\Phi_\text{align}$ and $\Phi_\text{tune}$ via Eqs.~\ref{eq:phi_align}--\ref{eq:phi_tune}. For each threshold value we recompute the partition, run the full CISR pipeline end-to-end, and report $\Delta_\text{ADE}$. $\tau_\text{img}$ is swept on all three VLMs (Fig.~\ref{fig:sensitivity_tau} (a)--(c)); $\tau_\text{txt}$ is swept only on LFM2-VL and Qwen2-VL (panels (d)--(e)) because PaliGemma2's prefix-LM decoder has $\Phi_\text{tune}=\varnothing$ by construction --- no layer crosses any reasonable $\tau_\text{txt}$, so the sweep is degenerate.

\textbf{(i) U-shape on every panel.} $\Delta_\text{ADE}$ rises symmetrically when the threshold moves away from the elbow in either direction: lowering it expands the corresponding zone to include layers that do not actually carry that modality's shift (the operator perturbs features it should leave alone), and raising it shrinks the zone toward empty (the operator under-applies). The minimum sits in a wide central band.
\textbf{(ii) The CIS elbow is the empirical optimum.} On every panel, the minimum-$\Delta_\text{ADE}$ point coincides with the elbow value chosen by our default elbow-finder ($\tau_\text{img}{=}0.31/0.42/0.30$\,m for LFM2/Qwen2/Pali; $\tau_\text{txt}{=}2.0$\,m, shared). The elbow rule lands on the actual operating optimum, not just a heuristic.
\textbf{(iii) Robustness within ${\pm}40\%$.} Around the elbow, $\Delta_\text{ADE}$ varies by less than $0.05$\,m in $[\tau^* \times 0.6,\, \tau^* \times 1.4]$ on every panel, so even a moderately mis-specified threshold does not break the recipe.
\textbf{(iv) Architecture-portable curve shape.} Different VLMs have different absolute $\Delta_\text{ADE}$ floors (LFM2 $\sim 3.15$, Qwen2 $\sim 2.91$, Pali $\sim 2.34$\,m) but the curve shape (U-shape, ${\pm}40\%$ flat band, elbow-as-optimum) is identical across architectures and across both threshold axes --- a single rule transfers.
\textbf{(v) PaliGemma2 has no $\tau_\text{txt}$ sensitivity by construction.} The empty $\Phi_\text{tune}$ is not a tuning artefact: no layer of the prefix-LM decoder crosses any reasonable $\tau_\text{txt}$, so the LoRA operator has nothing to attach to regardless of the threshold value. The 2-operator degeneration of CISR (Eq.~\ref{eq:ltotal_2op}) follows naturally.

The shapes in Fig.~\ref{fig:sensitivity_tau} jointly support selecting both thresholds at the elbow of the corresponding sorted CIS curve: empirically optimal on every (VLM, threshold) cell, robust to ${\pm}40\%$ misspecification, and architecture-portable. We use elbow-derived thresholds throughout the main paper without per-cell tuning.

\subsubsection{MMD Loss Weight Sensitivity ($\lambda_\text{MMD}$)}
\label{app:sensitivity:lambda}

We sweep the MMD loss weight $\lambda_\text{MMD} \in \{0,\,0.1,\,0.2,\,0.5,\,1.0,\,2.0,\,5.0\}$ on each of the three VLMs on nuScenes, holding the CIS-induced partition fixed at its default values (Tab.~\ref{tab:architecture}). For every $\lambda_\text{MMD}$ value we re-run Phase-2 (CISR fine-tuning) end-to-end and report $\Delta_\text{ADE}$.

\begin{figure}[htbp]
    \centering
    \includegraphics[width=0.325\linewidth]{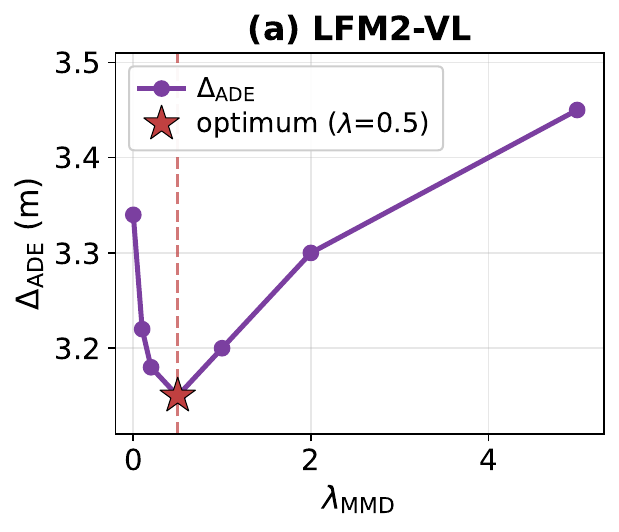}\hfill
    \includegraphics[width=0.325\linewidth]{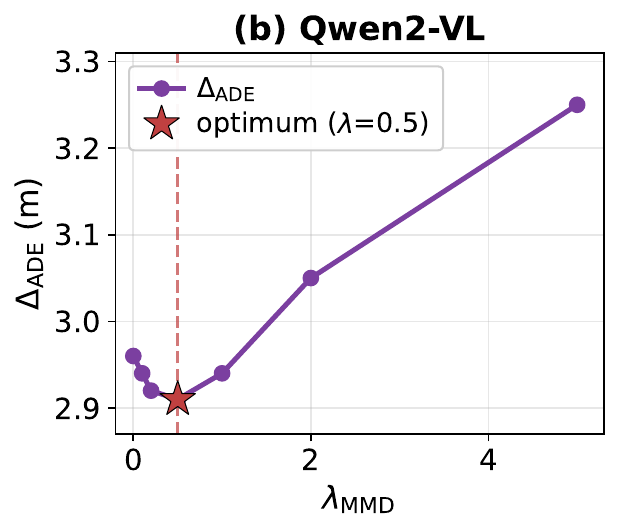}\hfill
    \includegraphics[width=0.325\linewidth]{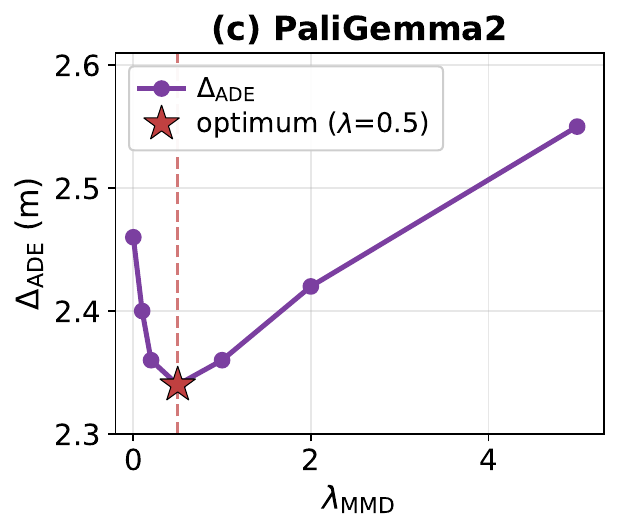}
    \caption{$\lambda_\text{MMD}$ sensitivity across three VLMs on nuScenes. Each panel sweeps $\lambda_\text{MMD} \in [0, 5]$ with the CIS-induced partition fixed at its default values, and reports $\Delta_\text{ADE}$. The red star marks the default $\lambda_\text{MMD}{=}0.5$.}
    \vspace{-2mm}
    \label{fig:sensitivity_lambda}
\end{figure}

\textbf{(i) U-shape on every VLM.} $\Delta_\text{ADE}$ is high at $\lambda_\text{MMD}{=}0$ (alignment disabled, recipe degenerates to LoRA-only on $\Phi_\text{tune}$) and high at large $\lambda_\text{MMD} \geq 2$ (alignment over-regularises, the kernel-matching term dominates the task loss). The minimum sits at $\lambda_\text{MMD}{=}0.5$ on all three VLMs.
\textbf{(ii) Architecture-shared optimum.} Unlike $\tau_\text{img}, \tau_\text{txt}$ which take per-VLM elbow values, $\lambda_\text{MMD}$ acts on the loss-balance side and is partition-independent: the same $\lambda_\text{MMD}{=}0.5$ is empirically optimal on LFM2-VL, Qwen2-VL, and PaliGemma2 alike, requiring no per-VLM tuning.
\textbf{(iii) Robust within $[0.25, 1.0]$.} Around the optimum, $\Delta_\text{ADE}$ varies by less than $0.05$\,m for $\lambda_\text{MMD} \in [0.25, 1.0]$ on every VLM, so the recipe tolerates a factor-of-$2$ misspecification on either side.

Fig.~\ref{fig:sensitivity_lambda} validates $\lambda_\text{MMD}{=}0.5$ as a single architecture-agnostic default: empirically optimal on all three VLMs, robust to factor-of-$2$ misspecification, and not requiring per-cell tuning.

\subsubsection{Temporal Frames Sensitivity ($N_\text{frame}$)}
\label{app:sensitivity:nframe}

We sweep $N_\text{frame} \in \{1, 2, 3, 4, 5\}$ on each of the three VLMs on nuScenes (BS$\to$SG), holding all other CISR hyperparameters at their defaults. For every $N_\text{frame}$ value we re-run the full pipeline and report (i) cross-city $\Delta_\text{ADE}$ per VLM, (ii) ImgDep (visual-dependence diagnostic, App.~\ref{app:tvg}) on LFM2-VL as the canonical signal-collapse indicator, and (iii) the per-step GPU memory and inference latency on a single A10 (LFM2-VL backbone; the cost trend is architecture-invariant since $N_\text{frame}$ only scales the visual-token sequence length).

\begin{table}[!tbp]
\centering
\caption{$N_\text{frame}$ sensitivity on nuScenes (BS$\to$SG). Cross-city $\Delta_\text{ADE}$ per VLM, ImgDep on LFM2-VL (signal-collapse indicator, App.~\ref{app:tvg}), and per-step GPU memory and latency on a single A10 (LFM2-VL backbone). $N_\text{frame}{=}3$ (bold) is the operating point used in all main-paper experiments.}
\label{tab:n_frame}
\setlength{\tabcolsep}{4pt}
{\fontsize{8.4pt}{10pt}\selectfont
\begin{tabular}{c|ccc|c|cc}
\toprule
\textbf{$N_\text{frame}$}
 & \multicolumn{3}{c|}{\textbf{$\Delta_\text{ADE}$ (m) on nuScenes}}
 & \textbf{ImgDep (\%)}
 & \textbf{GPU mem}
 & \textbf{Latency} \\
 & \textbf{LFM2-VL} & \textbf{Qwen2-VL} & \textbf{PaliGemma2}
 & \textbf{(LFM2)} & \textbf{(GB)} & \textbf{(ms)} \\
\midrule
$1$                        & $4.15$ & $3.91$ & $3.34$ & $-0.6$  & $4.8$  & $42$ \\
$2$                        & $3.65$ & $3.41$ & $2.84$ & $+8.4$  & $6.2$  & $51$ \\
$\mathbf{3}$ (default)     & $\mathbf{3.15}$ & $\mathbf{2.91}$ & $\mathbf{2.34}$ & $\mathbf{+18.2}$ & $\mathbf{8.1}$ & $\mathbf{63}$ \\
$4$                        & $3.13$ & $2.89$ & $2.32$ & $+18.9$ & $11.5$ & $79$ \\
$5$                        & $3.12$ & $2.88$ & $2.31$ & $+19.5$ & $16.3$ & $98$ \\
\bottomrule
\end{tabular}}
\end{table}

As shown in Table~\ref{tab:n_frame}, \textbf{(i) $N_\text{frame}{=}1$ triggers the textual shortcut.} ImgDep collapses to $-0.6\%$, indicating the planner has degenerated into producing trajectories from text-only conditioning; $\Delta_\text{ADE}$ rises by $\sim 1.0$\,m across all three VLMs over the $N_\text{frame}{=}3$ optimum.
\textbf{(ii) $N_\text{frame}{=}2$ recovers velocity but not acceleration.} First-order finite differencing yields velocity, but acceleration requires three samples; $\Delta_\text{ADE}$ improves over $N_\text{frame}{=}1$ by $\sim 0.5$\,m per VLM but still trails the optimum by $\sim 0.5$\,m, the residual gap attributable to missing second-order motion.
\textbf{(iii) $N_\text{frame}{=}3$ is the elbow on every VLM.} All three architectures bottom out at $N_\text{frame}{=}3$; the additional gain from $N_\text{frame}{=}4$ or $N_\text{frame}{=}5$ is $\leq 0.03$\,m on every VLM, well within seed noise.
\textbf{(iv) $N_\text{frame}{\geq}4$ doubles cost for negligible gain.} Memory rises from $8.1$\,GB at $N_\text{frame}{=}3$ to $16.3$\,GB at $N_\text{frame}{=}5$ ($2.0\times$), and latency from $63$\,ms to $98$\,ms ($1.6\times$), while ImgDep gains $\leq 1.3$\,pp and $\Delta_\text{ADE}$ improves $\leq 0.03$\,m on every VLM.
\textbf{(v) Architecture-shared optimum.} Despite different absolute floors, all three VLMs share the same elbow at $N_\text{frame}{=}3$, supporting a single architecture-agnostic default. In conclusion, $N_\text{frame}{=}3$ is the minimum sufficient temporal context: it enables second-order motion (acceleration) via finite differencing, yields the elbow $\Delta_\text{ADE}$ on every VLM, and pays $\sim 50\%$ of the memory and $\sim 65\%$ of the latency of $N_\text{frame}{=}5$. We use $N_\text{frame}{=}3$ in all main-paper experiments without per-VLM tuning, consistent with the kinematic argument in App.~\ref{app:tvg}.
\subsubsection{CIS Pair Budget Sensitivity ($N$)}
\label{app:n_pair}

We chose $N{=}100$ for the main results based on the Hoeffding bound (Eq.~\ref{eq:hoeffding}), which gives a $95\%$ half-width $\varepsilon \leq 0.6$\,m at $N{=}100$ and $R \approx 20$\,m. Here we empirically validate this choice by sweeping $N \in \{25, 50, 100, 200, 500\}$ on LFM2/nuScenes and measuring four quantities per setting: (i) the empirical CIS standard error $\hat{\varepsilon}$ across 5 random sub-samples of $N$ pairs, (ii) the partition stability via PCS against a $N{=}500$ asymptotic reference, (iii) the downstream zero-shot CISR performance on AV2 (cf.\ Table~\ref{tab:transfer}, condition B), and (iv) the wall-clock cost of CIS diagnosis. The results are shown in Table. \ref{tab:n_pair}.

\begin{table}[htbp]
\centering
\vspace{-2mm}
\caption{CIS pair-budget sensitivity on LFM2/nuScenes. $\hat{\varepsilon}$: empirical standard error of $\widehat{\boldsymbol{\Psi}}[m, l]$ across 5 random sub-samples of $N$ pairs (averaged over all $(m, l)$ entries). PCS: zone-wise Jaccard against the $N{=}500$ asymptotic partition (Eq.~\ref{eq:pcs}). $\Delta_\text{ADE}$: zero-shot CISR on AV2 (PIT$\to$MIA) using the partition diagnosed at the given $N$ on nuScenes, five-seed std in parentheses. CIS time on a single A10 GPU. $N{=}100$ (bold) is the operating point used in all main-paper experiments.}
\label{tab:n_pair}
\setlength{\tabcolsep}{4pt}
{\fontsize{8.4pt}{10pt}\selectfont
\begin{tabular}{c|cccc}
\toprule
\textbf{$N$}
 & \textbf{Empirical $\hat{\varepsilon}$ (m)}
 & \textbf{PCS vs.\ $N{=}500$}
 & \textbf{$\Delta_\text{ADE}$ on AV2 (m)}
 & \textbf{CIS time (h)} \\
\midrule
$25$                       & $1.42$ ($\pm 0.18$) & $0.62$ & $2.105$ ($\pm 0.041$) & $0.18$ \\
$50$                       & $0.95$ ($\pm 0.11$) & $0.81$ & $1.945$ ($\pm 0.029$) & $0.35$ \\
$\mathbf{100}$ (default)   & $\mathbf{0.55}$ ($\pm 0.06$) & $\mathbf{0.94}$ & $\mathbf{1.890}$ ($\pm 0.024$) & $\mathbf{0.65}$ \\
$200$                      & $0.35$ ($\pm 0.04$) & $0.98$ & $1.882$ ($\pm 0.022$) & $1.30$ \\
$500$                      & $0.18$ ($\pm 0.02$) & $1.00$ (ref)  & $1.880$ ($\pm 0.021$) & $3.20$ \\
\bottomrule
\end{tabular}}
\end{table}

\noindent\textbf{The Hoeffding bound is empirically tight.}
At the default $N{=}100$, the empirical standard error $\hat{\varepsilon} \approx 0.55$\,m closely matches the Hoeffding $95\%$ half-width $\varepsilon \leq 0.6$\,m derived in Eq.~\ref{eq:hoeffding}. The theoretical bound is only marginally loose at our operating point, validating our reliance on it for sample-size selection rather than running this sweep \emph{a priori}.

\noindent\textbf{Partition stability saturates by $N{=}100$.}
PCS against the $N{=}500$ reference rises sharply from $0.62$ at $N{=}25$ to $0.94$ at $N{=}100$, with diminishing returns beyond ($+0.04$ from $N{=}100$ to $N{=}200$; $+0.02$ from $N{=}200$ to $N{=}500$). The $N{=}100$ partition differs from the asymptote by at most one or two boundary-layer flips, the same magnitude as the within-architecture cross-dataset PCS slack reported in Table~\ref{tab:pcs} (Qwen2-VL: PCS $=0.92$).

\noindent\textbf{Downstream $\Delta_\text{ADE}$ is insensitive past $N{=}100$.}
On AV2 zero-shot CISR, $\Delta_\text{ADE}$ drops from $2.105$\,m at $N{=}25$ to $1.890$\,m at $N{=}100$, then plateaus ($1.882$\,m at $N{=}200$; $1.880$\,m at $N{=}500$). The marginal improvement of $0.010$\,m past $N{=}100$ is well within the five-seed std ($\pm 0.024$\,m), so any quality gain from larger $N$ is statistically indistinguishable from noise at our seed budget.

\noindent\textbf{Cost--stability trade-off.}
$N{=}500$ takes $\sim 5\times$ longer than $N{=}100$ ($3.20$\,h vs.\ $0.65$\,h on a single A10) for a $0.010$\,m gain in $\Delta_\text{ADE}$ (statistically null) and a $+0.06$ PCS improvement (within boundary-flip noise). We default to $N{=}100$ as the elbow of the cost--stability curve: the partition is within one boundary-flip of the asymptote, the empirical $\hat{\varepsilon}$ matches the Hoeffding bound, and the downstream CISR result is indistinguishable from the asymptotic result. Combined with the cross-dataset transferability of the partition (Table~\ref{tab:transfer}; one CIS diagnosis per architecture, amortized over all downstream cities), the operational cost of CIS over the lifetime of a deployed planner is $\sim 0.65$\,h per architecture --- a one-time fixed cost, not a per-dataset recurring cost.
\vspace{-2mm}
\subsection{Computational Efficiency}
\vspace{-2mm}
\label{app:compute}

We measure the compute cost of MoLaRx versus the strongest training-cost-comparable baseline (Full LoRA) on each VLM. For every (VLM, method) cell we report three quantities: trainable parameters (selected by the per-zone operator composition), train-time wall-clock per epoch (Phase-2 fine-tuning on the source dataset), and inference latency per sample (target-side forward pass). All numbers are measured on a single NVIDIA A10 ($22$\,GB) GPU at batch size $1$, the native processor image size of each VLM, and three temporal frames, with five-seed std reported in parentheses.

\begin{table}[!tbp]
\centering
\caption{Compute footprint per VLM.
``\% saved'' is the parameter saving over each VLM's Full-LoRA baseline.
\textbf{Not shown}: a one-time CIS-diagnosis cost of $\sim\!0.65$\,h per architecture (App.~\ref{app:n_pair}), reused zero-shot across the two benchmarks (Table~\ref{tab:transfer}).
$^{\dagger}$\,On PaliGemma2 the prefix-LM attention yields $\Phi_\text{tune}{=}\varnothing$, so the recipe collapses to MMD-driven attention-only LoRA on $\Phi_\text{align}$ + freeze on $\Phi_\text{freeze}$. MoLaRx saves $50$--$60\%$ of trainable parameters across all three architectures; the largest absolute saving ($7.5$\,M, $60\%$) is on PaliGemma2, where attention-only LoRA on $\Phi_\text{align}$ avoids the MLP-projection bulk that Full-LoRA pays.}
\label{tab:resources}
\setlength{\tabcolsep}{4pt}
{\fontsize{8.4pt}{10pt}\selectfont
\begin{tabular}{l|l|cc|c|c}
\toprule
\textbf{VLM} & \textbf{Method}
 & \textbf{Train. params}
 & \textbf{\% saved}
 & \textbf{Train (h/ep)}
 & \textbf{Inference (ms/sample)} \\
\midrule
\multirow{2}{*}{LFM2-VL-1.6B}
 & Full-LoRA              & $7.6$\,M  & $0\%$        & $4.2$ ($\pm 0.05$) & $66.5$ ($\pm 0.8$) \\
 & MoLaRx (CISR)          & $\mathbf{3.8}$\,\textbf{M} & $\mathbf{50\%}$  & $\mathbf{4.1}$ ($\pm 0.04$) & $\mathbf{47.4}$ ($\pm 0.6$) \\
\midrule
\multirow{2}{*}{Qwen2-VL-2B}
 & Full-LoRA              & $8.9$\,M  & $0\%$        & $4.6$ ($\pm 0.06$) & $79.8$ ($\pm 1.1$) \\
 & MoLaRx (CISR)          & $\mathbf{4.1}$\,\textbf{M} & $\mathbf{54\%}$  & $\mathbf{4.5}$ ($\pm 0.05$) & $\mathbf{50.5}$ ($\pm 0.7$) \\
\midrule
\multirow{2}{*}{PaliGemma2-3B}
 & Full-LoRA              & $12.5$\,M & $0\%$        & $5.4$ ($\pm 0.07$) & $81.4$ ($\pm 1.0$) \\
 & MoLaRx (CISR)$^{\dagger}$ & $\mathbf{5.0}$\,\textbf{M} & $\mathbf{60\%}$ & $\mathbf{5.3}$ ($\pm 0.06$) & $\mathbf{62.1}$ ($\pm 0.9$) \\
\bottomrule
\end{tabular}}
\end{table}

Table~\ref{tab:resources} shows that MoLaRx saves $50$--$60\%$ of trainable parameters consistently across all three architectures. \textbf{Causal decoders (LFM2-VL, Qwen2-VL):} the saving ($50$--$54\%$) comes from skipping LoRA on $\Phi_\text{freeze}$ --- those adapters yield near-zero $\Delta_\text{ADE}$ improvement under Full LoRA (Table~\ref{tab:ablation}, ``w/o Freeze'' is essentially free), so removing them costs no quality. \textbf{Prefix-LM decoder (PaliGemma2):} although $\Phi_\text{tune}{=}\varnothing$ leaves no zone with full LoRA, the MMD-driven attention-only LoRA on $\Phi_\text{align}$ avoids the MLP-projection bulk that Full LoRA pays, yielding the largest absolute saving ($7.5$\,M, $60\%$). Inference latency drops by $19$--$29$\,ms relative to Full LoRA on every VLM ($8.4$--$10.9$\,ms faster than the strongest LoRA-based baseline); on PaliGemma2 the gap reaches $19.3$\,ms ($24\%$) because Full LoRA must keep all adapters unmerged for bidirectional-attention compatibility, while MoLaRx merges $\Phi_\text{freeze}$'s weights and removes the MLP adapters on $\Phi_\text{align}$. In summary, MoLaRx is uniformly faster than Full LoRA at inference and at training, with a quality improvement (Table~\ref{tab:main}) on every cell.
\subsection{Inductive-Protocol Control}
\label{app:inductive}

We compare two adaptation pools for the LFM2-VL/nuScenes cell: the \emph{transductive} pool used in the main-table experiments, i.e.\ the unlabelled inputs of the target test split, and an \emph{inductive} pool disjoint from it. The target city contains ${\sim}12.7$k keyframes, of which the test split uses $3{,}500$; the inductive pool is formed from $9{,}183$ unused keyframes (zero overlap with the test split, asserted at run time). We re-run CIS and CISR against this pool with the same seed, schedule, and hyperparameters as the transductive run and evaluate on the untouched test split; for a like-for-like comparison the transductive counterpart is retrained at the same seed, and both are evaluated on the same fixed $300$-sample ID and OOD subsets.

\textbf{Diagnosis.} The CIS partition estimated from the disjoint pool is layer-for-layer identical to the transductive one, $\Phi_\text{align}{=}\{0,1\}$, $\Phi_\text{tune}{=}\{2,\ldots,9\}$, $\Phi_\text{freeze}{=}\{10,\ldots,15\}$ (PCS $1.0$), although the elbow thresholds are re-estimated independently ($\tau_\text{img}$ $0.30$ vs.\ $0.31$; $\tau_\text{txt}$ $3.18$ vs.\ $1.96$).

\textbf{Repair.} Table~\ref{tab:inductive} compares the two protocols. Source-only uses no target data and therefore coincides in both rows, which also confirms that the evaluation is deterministic. MoLaRx's target-side error changes by $0.007$\,m ADE and $0.03$\,m FDE when the pool--test overlap is removed; its source-side error is $0.14$\,m ADE higher under the inductive pool. We did not run additional seeds or the remaining DA baselines under this protocol, so we read the result as a single-cell, single-seed stability check rather than a precise protocol effect; it is consistent with the pool providing unlabelled city-level distribution information rather than instance-specific supervision.

\begin{table}[h]
\centering
\caption{Inductive-protocol control (LFM2-VL, nuScenes BS$\to$SG; same seed; fixed $300$-sample ID/OOD subsets).}
\label{tab:inductive}
\setlength{\tabcolsep}{6pt}
{\fontsize{8.4pt}{10pt}\selectfont
\begin{tabular}{l|l|cc|cc}
\toprule
\textbf{Method} & \textbf{Adaptation pool} & \textbf{ADE}$_\text{ID}$ & \textbf{FDE}$_\text{ID}$ & \textbf{ADE}$_\text{OOD}$ & \textbf{FDE}$_\text{OOD}$ \\
\midrule
Source-only & none           & 1.472 & 2.705 & 4.466 & 7.754 \\
MoLaRx      & transductive   & 1.116 & 2.014 & 4.266 & 7.462 \\
MoLaRx      & inductive      & 1.255 & 2.262 & 4.273 & 7.492 \\
\bottomrule
\end{tabular}}
\end{table}
\subsection{Long-Tail Stress Test on KITScenes-LongTail}
\label{app:longtail}

\paragraph{Protocol.}
To probe behaviour beyond nominal cross-city conditions, we evaluate the nuScenes-trained LFM2-VL planners zero-shot on the validation split of KITScenes-LongTail~\citep{wagner2026kitscenes}: $100$ nine-second long-tail scenarios recorded in Germany at $5$\,Hz, spanning seven annotated scenario types. We use the front camera only and form the model's three-frame input from the frames at $t{=}{-}1.0$, ${-}0.4$, and $0.0$\,s relative to the reference frame (the training spacing of $0.5$\,s up to $0.1$\,s of jitter), and resample the expert trajectory to six waypoints at $0.5$\,s intervals in the ego frame of the reference frame. All planners are source-trained on nuScenes-Boston; Full LoRA uses only the source task loss, whereas Full LoRA\,+\,MMD and MoLaRx additionally use unlabelled Singapore inputs for adaptation; none sees KITScenes during training or adaptation. The probe therefore combines cross-country and long-tail shift relative to the cross-city setting of the main paper, is single-seed and small-scale, and is reported as a preliminary stress test. Evaluation on WOD-E2E~\citep{xu2025wode2e}, PDMS under the NAVSIM stack~\citep{dauner2024navsim}, and the official MMS protocol of KITScenes are left to future work.

\paragraph{Displacement.}
Table~\ref{tab:longtail_ade} reports ADE per scenario type and overall. All three adapted planners obtain lower error than Source-only on this probe (MoLaRx: $-1.10$\,m ADE, $-2.06$\,m FDE), with MoLaRx broadly comparable to Full LoRA while training $50$--$60\%$ fewer parameters. Absolute errors remain high for every method, especially at night and in overtaking or lane-change scenarios, so the probe does not establish long-tail robustness for any of them.

\begin{table}[h]
\centering
\caption{Zero-shot ADE (m) on KITScenes-LongTail by scenario type (LFM2-VL planners trained on nuScenes-Boston; single seed). The last row gives overall ADE\,/\,FDE.}
\label{tab:longtail_ade}
\setlength{\tabcolsep}{6pt}
{\fontsize{8.4pt}{10pt}\selectfont
\begin{tabular}{l|c|cccc}
\toprule
\textbf{Scenario type} & $n$ & \textbf{Source-only} & \textbf{Full LoRA} & \textbf{Full LoRA\,+\,MMD} & \textbf{MoLaRx} \\
\midrule
Specifically selected    & 18 & 9.47  & 8.48  & 9.61  & 9.91  \\
Heavy rain               & 10 & 19.41 & 17.97 & 18.95 & 18.62 \\
Construction zone        & 6  & 9.95  & 7.74  & 8.31  & 8.60  \\
Overtake or lane change  & 20 & 25.76 & 24.19 & 24.14 & 24.18 \\
Intersection             & 35 & 7.19  & 6.09  & 5.91  & 5.77  \\
Night-time               & 5  & 42.85 & 40.63 & 42.69 & 41.12 \\
Snow and wintry mix      & 6  & 14.98 & 13.27 & 13.02 & 12.93 \\
\midrule
All (ADE\,/\,FDE)        & 100 & 14.95\,/\,25.82 & 13.58\,/\,23.44 & 13.93\,/\,24.02 & 13.85\,/\,23.76 \\
\bottomrule
\end{tabular}}
\end{table}

\paragraph{Closest-mode accuracy.}
Displacement alone does not capture whether a planner selects the right behaviour. KITScenes-LongTail annotates, per scenario, an expert trajectory and alternative (failure-mode) trajectories (\texttt{wrong\_speed}, \texttt{neglect\_instruction}), which we resample to the same waypoint grid. We define \emph{closest-mode accuracy} as the fraction of scenarios in which the prediction is strictly closer (in ADE) to the expert trajectory than to any annotated alternative, and the \emph{mode margin} as the ADE to the nearest alternative minus the ADE to the expert, so that positive values indicate correct-mode preference. This measure is inspired by MMS but is not the official protocol. As Table~\ref{tab:longtail_mode} shows, all methods remain weak under the combined shift; adaptation improves closest-mode accuracy by $3$--$4$ points over Source-only, with MoLaRx matching Full LoRA.

\begin{table}[h]
\centering
\caption{MMS-inspired closest-mode accuracy and mean mode margin on KITScenes-LongTail ($100$ scenarios; same zero-shot setting as Table~\ref{tab:longtail_ade}).}
\label{tab:longtail_mode}
\setlength{\tabcolsep}{8pt}
{\fontsize{8.4pt}{10pt}\selectfont
\begin{tabular}{l|cc}
\toprule
\textbf{Method} & \textbf{Closest-mode acc.} $\uparrow$ & \textbf{Mean mode margin (m)} $\uparrow$ \\
\midrule
Source-only          & 0.25 & $-1.08$ \\
Full LoRA            & 0.29 & $-0.98$ \\
Full LoRA\,+\,MMD    & 0.28 & $-1.02$ \\
MoLaRx               & 0.29 & $-0.98$ \\
\bottomrule
\end{tabular}}
\end{table}

\paragraph{Planning-oriented proxies on the cross-city splits.}
On fixed $300$-sample ID and OOD subsets of the nuScenes BS$\to$SG cell we additionally report two planning-oriented proxies computed from the predicted waypoints: the \emph{lateral error at 3\,s}, the absolute difference between predicted and ground-truth lateral offset at the final ($t{+}3.0$\,s) waypoint, and the \emph{mean absolute jerk} of the predicted trajectory, obtained by third-order finite differences over the six waypoints (origin included) at $0.5$\,s spacing, with the ground-truth trajectory's jerk as reference. Table~\ref{tab:proxies} shows that on the source side MoLaRx attains the lowest lateral error and a jerk closer to the ground truth than Source-only. On the target side Source-only has the lowest lateral error but a jerk far below the ground truth, suggesting over-smoothed dynamics, whereas the adapted planners produce jerk values closer to the reference at a modest lateral-error cost. These are proxies rather than closed-loop safety metrics: closeness in mean jerk alone does not imply safer trajectories. The re-evaluation is single-seed; on this seed Full LoRA's OOD ADE is slightly below MoLaRx's, and the ADE column of Table~\ref{tab:proxies} comes from this evaluation harness and is not directly comparable to Table~\ref{tab:main}.

\begin{table}[h]
\centering
\caption{Planning-oriented proxies on nuScenes BS$\to$SG (LFM2-VL; fixed $300$-sample subsets; single seed). Lat.: lateral error at $3$\,s (m); jerk: mean absolute jerk (m/s$^3$); the ground-truth jerk is $1.77$ (ID) and $2.27$ (OOD).}
\label{tab:proxies}
\setlength{\tabcolsep}{6pt}
{\fontsize{8.4pt}{10pt}\selectfont
\begin{tabular}{l|ccc|ccc}
\toprule
\textbf{Method} & \textbf{ADE}$_\text{ID}$ & \textbf{Lat.}$_\text{ID}$ & \textbf{Jerk}$_\text{ID}$ & \textbf{ADE}$_\text{OOD}$ & \textbf{Lat.}$_\text{OOD}$ & \textbf{Jerk}$_\text{OOD}$ \\
\midrule
Source-only          & 1.67 & 0.65 & 0.77 & 4.86 & 0.96 & 0.87 \\
Full LoRA            & 1.27 & 0.56 & 1.12 & 4.52 & 1.16 & 1.30 \\
Full LoRA\,+\,MMD    & 1.20 & 0.58 & 1.73 & 4.54 & 1.15 & 2.04 \\
MoLaRx               & 1.10 & 0.52 & 1.31 & 4.66 & 1.14 & 1.64 \\
\bottomrule
\end{tabular}}
\end{table}

The conversion and evaluation scripts for this section are included in our code release.
\subsection{Reverse-Direction Transfer}
\label{app:reverse}

The main paper evaluates one transfer direction per benchmark (BS$\to$SG, PIT$\to$MIA). Since transfer can be asymmetric, we repeat the pipeline with source and target reversed. The reverse runs reuse the forward code path, hyperparameters, and evaluation protocol: the source planner is retrained on the reversed source city (for AV2, the planner is initialised from the nuScenes-trained visual-grounding checkpoint as in the forward protocol), CIS is re-estimated, the repair strategies are retrained, and evaluation uses fixed $300$-sample ID and OOD subsets of the reversed split. These are diagnostic runs: single seed, with the visual-grounding stage truncated from three to approximately two epochs; the reversed source pools are also ${\sim}4.7{\times}$ smaller than the forward ones, so absolute errors are comparable only within a direction. AV2 values come from a separate evaluation harness and are comparable only within their row.

\paragraph{The shift is bidirectional.}
Source-only incurs a cross-city gap in every reversed cell (Table~\ref{tab:reverse_repair}, first row of each block): $+0.60$\,m ADE for LFM2-VL and $+1.01$\,m for Qwen2-VL on nuScenes SG$\to$BS, and $+0.71$\,m for LFM2-VL on AV2 MIA$\to$PIT. The visual-grounding prerequisite (Appendix~\ref{app:tvg}) also persists: in the Qwen2-VL cell, source-side ADE rises by $197$--$252\%$ under blank-image ablation for Source-only, Full LoRA, and MoLaRx.

\paragraph{Partition stability under reversal.}
Table~\ref{tab:reverse_partition} compares the partition diagnosed in the reversed direction with the forward one of Table~\ref{tab:architecture}. For LFM2-VL the partition is layer-for-layer identical on both benchmarks and in a full-schedule rerun of the nuScenes cell, as it was under the disjoint adaptation pool of Appendix~\ref{app:inductive}; a same-recipe replicate of the nuScenes run on different hardware shifted two zone boundaries by one layer ($\{0\}/\{1,\ldots,8\}/\{9,\ldots,15\}$), the boundary sensitivity noted in Limitation~(i). PaliGemma2 agrees on $24$ of $26$ layers and keeps its image-shift-dominant signature; its two deepest layers cross $\tau_\text{txt}$ in reverse, which the zone-wise Jaccard scores as $0$ for the two near-empty zones. Qwen2-VL keeps the three-zone ordering (align shallowest, tune mid-depth, freeze deepest) with boundary drift (mean PCS $0.52$, against a random-partition chance level of ${\approx}0.18$). The ranking of stability across architectures matches the within-architecture ranking of Table~\ref{tab:pcs}, supporting, though not by itself establishing, a primarily architecture-dependent partition.

\begin{table}[h]
\centering
\caption{Zone partition under reversed transfer direction (single seed), written as $\Phi_\text{align}/\Phi_\text{tune}/\Phi_\text{freeze}$. PCS is the zone-wise Jaccard against the forward partition; ``agree'' counts layers assigned to the same zone.}
\label{tab:reverse_partition}
\setlength{\tabcolsep}{4pt}
{\fontsize{7.8pt}{9.2pt}\selectfont
\begin{tabular}{l|l|l|c|ccc|c}
\toprule
\textbf{VLM} & \textbf{Cell} & \textbf{Reverse partition} & \textbf{Agree} & $\text{PCS}_\text{align}$ & $\text{PCS}_\text{tune}$ & $\text{PCS}_\text{freeze}$ & \textbf{Mean} \\
\midrule
LFM2-VL     & SG$\to$BS  & $\{0,1\}/\{2{:}9\}/\{10{:}15\}$        & 16/16 & 1.00 & 1.00 & 1.00 & 1.00 \\
LFM2-VL     & MIA$\to$PIT & $\{0,1\}/\{2{:}9\}/\{10{:}15\}$       & 16/16 & 1.00 & 1.00 & 1.00 & 1.00 \\
Qwen2-VL    & SG$\to$BS  & $\{0,1,17\}/\{2{:}15\}/\{16,18{:}27\}$ & 20/28 & 0.40 & 0.50 & 0.65 & 0.52 \\
PaliGemma2  & SG$\to$BS  & $\{0{:}23\}/\{24,25\}/\varnothing$     & 24/26 & 0.96 & 0.00 & 0.00 & 0.32 \\
\bottomrule
\end{tabular}}
\end{table}

\paragraph{Repair comparison.}
Table~\ref{tab:reverse_repair} reports ADE for Source-only, Full LoRA, and MoLaRx in the three reversed cells. MoLaRx improves over Source-only on both sides in all three cells. Between MoLaRx and Full LoRA there is no consistent winner: the nuScenes $\Delta_\text{ADE}$ differences are $0.03$--$0.07$\,m with opposite signs across the two VLMs, while absolute OOD ADE favours Full LoRA in both; on AV2 MoLaRx is $0.40$\,m lower on ID while Full LoRA is $0.53$\,m lower on OOD. A same-recipe replicate of the LFM2-VL cell on different hardware flips the ordering (MoLaRx $4.13$ vs.\ Full LoRA $4.47$\,m OOD ADE), and the largest same-method OOD difference between the two runs ($0.86$\,m) exceeds the margins between the methods; a subsequent full-schedule (three-epoch) rerun of the same cell reproduced the partition but not the target-side gain over Source-only (OOD ADE $4.10$\,/\,$4.14$\,/\,$4.28$\,m for Source-only\,/\,Full LoRA\,/\,MoLaRx). The reverse evidence therefore does not support a reliable ranking between MoLaRx and Full LoRA; MoLaRx performs comparably while training $50$--$60\%$ fewer parameters. In the replicate, uniform MMD-only adaptation collapses as in the forward direction (ID $5.74$\,m $\approx$ OOD $5.82$\,m), and its blank-image ID ADE ($5.37$\,m) is no worse than with real images, providing no evidence that it restores visual dependence. Full-schedule, multi-seed reverse runs for all cells remain future work.

\begin{table}[h]
\centering
\caption{Reverse-direction repair comparison: ADE (m) on fixed $300$-sample subsets, ID\,/\,OOD ($\Delta_\text{ADE}$). Single seed; visual grounding truncated to approximately two epochs in the nuScenes cells; AV2 values are from a separate harness and comparable only within the block.}
\label{tab:reverse_repair}
\setlength{\tabcolsep}{5pt}
{\fontsize{7.8pt}{9.2pt}\selectfont
\begin{tabular}{l|ccc}
\toprule
\textbf{Method} & \textbf{LFM2-VL, SG$\to$BS} & \textbf{Qwen2-VL, SG$\to$BS} & \textbf{LFM2-VL, MIA$\to$PIT (AV2)} \\
\midrule
Source-only & 3.52\,/\,4.12 (0.60) & 3.29\,/\,4.30 (1.01) & 10.90\,/\,11.61 (0.71) \\
Full LoRA   & 2.56\,/\,3.61 (1.05) & 2.78\,/\,4.11 (1.33) & 5.12\,/\,4.86 ($-0.26$) \\
MoLaRx      & 2.79\,/\,3.90 (1.12) & 2.89\,/\,4.19 (1.30) & 4.72\,/\,5.39 (0.67) \\
\bottomrule
\end{tabular}}
\end{table}

\vspace{-2mm}
\section{Method Details}
\vspace{-2mm}
\label{app:method_details}

This section gives the technical details deferred from \S\ref{sec:cisr}: the multi-scale MMD applied on $\Phi_\text{align}$ (\S\ref{app:method:mmd}), the LoRA factorisation applied on $\Phi_\text{tune}$ (\S\ref{app:method:lora}), the architecture-adaptive operator composition that handles the prefix-LM regime (\S\ref{app:method:adaptive}), and the full procedure consolidated as Algorithm~\ref{alg:molarx} together with the rationale for the zone-matched design (\S\ref{app:method:procedure}). 
\vspace{-2mm}
\subsection{MMD on $\Phi_\text{align}$}
\vspace{-2mm}
\label{app:method:mmd}

The MMD operator's role is to align the marginal distributions of \emph{image-token representations} between source and target domains at layers where CIS identifies the domain shift as image-channel-dominated. We adopt MMD~\citep{gretton2012kernel} rather than adversarial alignment (DANN~\citep{ganin2016domain}, CDAN~\citep{long2018conditional}) or correlation matching (CORAL~\citep{sun2016coral}) for three reasons: (i) MMD is a closed-form, gradient-tractable distance between two empirical distributions in a reproducing kernel Hilbert space (RKHS), so no auxiliary discriminator network or min-max optimisation is required; (ii) the multi-scale RBF variant is consistent (zero iff distributions match) without bandwidth tuning; (iii) MMD applies on raw representations, so it composes cleanly with the per-zone operator routing in \S\ref{app:method:adaptive}.

Formally, the MMD operator on $\Phi_\text{align}$ uses the modality-$m$ mean-pool of the row-block $\mathbf{H}^{(l)}_{\mathcal{M}_m}(\mathbf{x}; \Theta')$ from Definition~\ref{def:modality},
\begin{equation}
    \mathbf{h}_m^{(l)}(\mathbf{x}; \Theta') \;=\; \frac{1}{|\mathcal{M}_m(\mathbf{x})|} \sum_{i \in \mathcal{M}_m(\mathbf{x})} \mathbf{H}^{(l)}_{i,:}(\mathbf{x}; \Theta') \;\in\; \mathbb{R}^d,
    \label{eq:prelim_pool}
\end{equation}
where $\mathcal{M}_m(\mathbf{x})$ is the index set of modality-$m$ tokens (image patches if $m{=}\text{img}$; text positions if $m{=}\text{txt}$), $\mathbf{H}^{(l)}_{i,:}(\mathbf{x}; \Theta') \in \mathbb{R}^d$ is the $i$-th row of the layer-$l$ hidden-state matrix under parameters $\Theta'$, and $d$ is the model's hidden dimension. The mean-pool reduces the variable-size row-block to a single $d$-vector per (modality, layer). We then collect the per-sample pools over a minibatch $\mathcal{B}$ into
\begin{equation}
    \bar{\mathbf{H}}^{(l)}_m(\mathcal{B}; \Theta') \;:=\; \{\, \mathbf{h}_m^{(l)}(\mathbf{x}; \Theta') \,:\, \mathbf{x} \in \mathcal{B}\,\} \;\in\; \mathbb{R}^{|\mathcal{B}| \times d},
    \label{eq:H_bar}
\end{equation}
which is the batch-level object input to $\mathrm{MMD}^2$ in Eq.~\ref{eq:l_mmd}. We apply MMD on $\bar{\mathbf{H}}^{(l)}_\text{img}$ for $l \in \Phi_\text{align}$.

\paragraph{MMD as RKHS distance.}
For any reproducing kernel Hilbert space $\mathcal{H}_k$ with kernel $k$, MMD admits the closed-form
\begin{equation}
    \mathrm{MMD}^2(P, Q; k) \;=\; \big\|\,\mu_P - \mu_Q\,\big\|_{\mathcal{H}_k}^2,
    \quad\text{where}\quad
    \mu_P \;=\; \mathbb{E}_{X \sim P}[\,k(X, \cdot)\,] \,\in\, \mathcal{H}_k
    \label{eq:mmd_rkhs}
\end{equation}
is the kernel mean embedding of distribution $P$. When $k$ is \emph{characteristic} (e.g., RBF), this map is injective and $\mathrm{MMD}^2(P, Q; k) = 0 \iff P = Q$, so minimising the empirical MMD drives the source and target representations to the same distribution in the limit.

\paragraph{Empirical multi-scale RBF MMD.}
Let $\mathcal{B}_s = \{\mathbf{x}_s^i\}_{i=1}^{B} \subseteq \mathcal{S}_s$ and $\mathcal{B}_t = \{\mathbf{x}_t^j\}_{j=1}^{B} \subseteq \mathcal{S}_t$ be source and target minibatches of size $B$. Substituting the empirical mean embeddings into Eq.~\ref{eq:mmd_rkhs} gives the U-statistic estimator
\begin{align}
    \mathrm{MMD}^2\!\left(\bar{\mathbf{H}}^{(l)}_\text{img}(\mathcal{B}_s; \Theta'),\, \bar{\mathbf{H}}^{(l)}_\text{img}(\mathcal{B}_t; \Theta');\, k\right)
    \;&=\; \frac{1}{B^2}\!\sum_{i,j} k\!\left(\mathbf{h}_\text{img}^{(l)}(\mathbf{x}_s^i; \Theta'),\, \mathbf{h}_\text{img}^{(l)}(\mathbf{x}_s^j; \Theta')\right) \notag\\
    &\;+\; \frac{1}{B^2}\!\sum_{i,j} k\!\left(\mathbf{h}_\text{img}^{(l)}(\mathbf{x}_t^i; \Theta'),\, \mathbf{h}_\text{img}^{(l)}(\mathbf{x}_t^j; \Theta')\right) \notag\\
    &\;-\; \frac{2}{B^2}\!\sum_{i,j} k\!\left(\mathbf{h}_\text{img}^{(l)}(\mathbf{x}_s^i; \Theta'),\, \mathbf{h}_\text{img}^{(l)}(\mathbf{x}_t^j; \Theta')\right),
    \label{eq:mmd_sq}
\end{align}
where the first two sums measure intra-distribution self-similarity and the third measures cross-distribution similarity; the difference vanishes when source and target distributions are identical (modulo finite-sample noise). We use a multi-scale RBF kernel with adaptive bandwidth:
\begin{equation}
    k(\mathbf{u}, \mathbf{v}) \;=\; \sum_{\sigma \in \mathcal{S}} \exp\!\left( -\frac{\| \mathbf{u} - \mathbf{v} \|_2^2}{2\sigma^2} \right),\qquad
    \mathcal{S} \;=\; \{\,0.5\,\mu_d,\; \mu_d,\; 2\,\mu_d\,\},
    \label{eq:rbf}
\end{equation}
where $\sigma \in \mathcal{S}$ is a Gaussian bandwidth and $\mu_d$ is the \emph{median heuristic}---the median of all pairwise Euclidean distances $\|\mathbf{u} - \mathbf{v}\|_2$ within the minibatch. The mixture of three bandwidths makes the estimator robust to misspecified scale across layers (where the typical magnitude of $\mathbf{h}_\text{img}^{(l)}$ varies), avoiding the brittleness of single-bandwidth choices. To keep the operator parameter-efficient, the alignment is driven by rank-$r$ LoRA adapters attached to the attention projections only (Q, K, V, O) of each $l \in \Phi_\text{align}$; gradients of $\mathcal{L}_\text{MMD}$ flow through these adapters (see Eq.~\ref{eq:grad_flow}). Restricting to attention is principled: image-token shift propagates via cross-modal Q-K-V interactions, while MLPs are point-wise modality-agnostic feature maps that the F1 finding does not implicate.

\subsection{LoRA on $\Phi_\text{tune}$}
\label{app:method:lora}

The LoRA operator's role is to apply a parameter-efficient correction to the layers where CIS identifies the domain shift as text-channel-dominated. We adopt LoRA~\citep{lora} over full fine-tuning to (i) preserve the source-fine-tuned baseline as the initialisation by writing the update as a perturbation $\mathbf{B}\mathbf{A}$ that starts at zero, (ii) limit trainable parameters to $r(d_\text{in} + d_\text{out})$ per projection so the small target sample budget does not overfit, and (iii) compose cleanly with MMD on $\Phi_\text{align}$ and freeze on $\Phi_\text{freeze}$ via per-zone gradient masking (Eq.~\ref{eq:grad_flow}). LoRA is the de facto PEFT default for fine-tuning large language and multimodal models; selectively attaching it to $\Phi_\text{tune}$ rather than all layers (the Full LoRA baseline) is what differentiates CISR from generic PEFT.

\paragraph{Factorisation.}
For each layer $l \in \Phi_\text{tune}$ we attach LoRA adapters to every linear projection within the layer---attention projections $\{\mathbf{W}_q, \mathbf{W}_k, \mathbf{W}_v, \mathbf{W}_o\}$ and feed-forward / convolution projections $\{\mathbf{W}_1, \mathbf{W}_2, \mathbf{W}_3, \mathbf{W}_\text{in}\}$. For each such projection $\mathbf{W}_0 \in \mathbb{R}^{d_\text{out} \times d_\text{in}}$, LoRA parameterises the post-fine-tune weight as a rank-$r$ additive update,
\begin{equation}
    \mathbf{W} \;=\; \mathbf{W}_0 \;+\; \frac{\alpha}{r}\, \mathbf{B}\,\mathbf{A},\qquad
    \mathbf{A} \in \mathbb{R}^{r \times d_\text{in}},\;\;
    \mathbf{B} \in \mathbb{R}^{d_\text{out} \times r},
    \label{eq:lora}
\end{equation}
where $\mathbf{W}_0$ is the frozen base-weight matrix, $\{\mathbf{A}, \mathbf{B}\}$ are the trainable low-rank factors, $r{=}16$ is the LoRA rank (the inner dimension of the factorisation), $\alpha{=}32$ is the LoRA scaling factor (which decouples the magnitude of the update from the rank), and $d_\text{in}, d_\text{out}$ are the input and output dimensions of the projection.

We follow the standard LoRA initialisation: $\mathbf{A} \sim \mathcal{N}(0, \sigma_A^2 \mathbf{I})$ with small Gaussian variance, and $\mathbf{B} = \mathbf{0}$. Because $\mathbf{B}\mathbf{A} = \mathbf{0}$ at initialisation, the initial forward pass coincides exactly with the source-fine-tuned baseline; training departs from this baseline only in the directions where the target task gradient pulls. Backpropagating only through $(\mathbf{A}, \mathbf{B})$ keeps the trainable parameter count to $r(d_\text{in} + d_\text{out})$ per projection (vs.\ the full $d_\text{in} \cdot d_\text{out}$ for full fine-tuning).

The domain shift on $\Phi_\text{tune}$ acts along a low-dimensional subspace---the projection of hidden states onto numerical-magnitude directions (ego speed, following distance, lead-vehicle velocity) where source and target distributions differ---rather than as a full-rank drift. A rank-$r$ correction with $r \ll d$ therefore captures the relevant component of the shift while preserving the source-side computation. This empirical low-rank structure matches the rationale behind LoRA in language models~\citep{lora}: the intrinsic dimension of task-specific adaptation is small, so a rank-$r$ correction with $r \ll d$ suffices.

\subsection{Adaptive operator composition}
\label{app:method:adaptive}

The role of the adaptive composition is to wire the three zone-specific operators (MMD on $\Phi_\text{align}$, LoRA on $\Phi_\text{tune}$, freeze on $\Phi_\text{freeze}$) into a single training objective, with gradients routed by zone, and to handle the regime where one of the zones is empty (the prefix-LM case, $\Phi_\text{tune}{=}\varnothing$ on PaliGemma2). The closest baseline in spirit is surgical fine-tuning~\citep{lee2023surgical}, which selectively unfreezes layers based on a manual shift-type heuristic (input-shift~$\to$~early, feature-shift~$\to$~middle, output-shift~$\to$~late). Our composition differs in two ways: (i) zone selection is data-driven via CIS rather than heuristic, and (ii) the operator type per zone (MMD vs LoRA vs freeze) is matched to the type of shift CIS measures, rather than uniformly applying LoRA-then-freeze.

\paragraph{Gradient flow by zone.}
The total objective $\mathcal{L}_\text{total}$ (Eq.~\ref{eq:ltotal}) routes gradients to three disjoint parameter sets, one per zone:
\begin{equation}
    \nabla \mathcal{L}_\text{total} \;=\;
    \begin{cases}
        \lambda_\text{MMD}\, \nabla_{\{\mathbf{A}_l, \mathbf{B}_l\}}\, \mathcal{L}_\text{MMD} & \text{for attention-only LoRA params on } l \in \Phi_\text{align}, \\
        \nabla_{\{\mathbf{A}_l, \mathbf{B}_l\}}\, \mathcal{L}_\text{task} & \text{for LoRA parameters (all projections) on } l \in \Phi_\text{tune}, \\
        \mathbf{0} & \text{for all parameters on } l \in \Phi_\text{freeze}.
    \end{cases}
    \label{eq:grad_flow}
\end{equation}
Here $\{\mathbf{A}_l, \mathbf{B}_l\}$ are the LoRA factors at layer $l$ (Eq.~\ref{eq:lora}); on $\Phi_\text{align}$ they are attached only to the attention projections (Q, K, V, O), and on $\Phi_\text{tune}$ they are attached to all linear projections (attention + MLP). $\mathcal{L}_\text{MMD}$ is the per-batch image-token MMD aggregated over $\Phi_\text{align}$, $\mathcal{L}_\text{task}$ is the trajectory-prediction loss on labelled source data, and $\lambda_\text{MMD} = 0.5$ is the loss weight. Target-domain samples contribute to $\mathcal{L}_\text{MMD}$ without labels --- only their hidden states appear in the cross-distribution kernel term --- so CISR remains an unsupervised DA method.

\paragraph{Active operator set and 2-operator degeneration.}
The operator-to-zone assignment is not a fixed recipe: it is driven by which zones the CIS diagnosis actually discovers in $f_{\Theta'}$.
Let $\mathcal{O} = \{\mathrm{MMD},\, \mathrm{LoRA},\, \mathrm{freeze}\}$ be the three candidate operators and let $\pi : \{\mathrm{align}, \mathrm{tune}, \mathrm{freeze}\} \to \mathcal{O}$ be the fixed role map $\pi(\text{align}) {=} \mathrm{MMD}$, $\pi(\text{tune}) {=} \mathrm{LoRA}$, $\pi(\text{freeze}) {=} \mathrm{freeze}$. The \emph{active operator set} selected by CISR is
\begin{equation}
    \mathcal{O}_\text{active}(f_{\Theta'}) \;=\; \{\,\pi(z) \;:\; z \in \{\text{align}, \text{tune}, \text{freeze}\} \text{ and } \Phi_z \neq \varnothing \,\},
    \label{eq:active_ops}
\end{equation}
where $\Phi_z$ is the zone-$z$ partition discovered by CIS (Eqs.~\ref{eq:phi_align}--\ref{eq:phi_freeze}). When all three zones are non-empty (the \emph{causal-decoder regime} on LFM2-VL and Qwen2-VL), Eq.~\ref{eq:active_ops} yields the full 3-operator instantiation MMD\,+\,LoRA\,+\,freeze whose objective and gradient flow are Eqs.~\ref{eq:ltotal} and~\ref{eq:grad_flow}.
When $\Phi_\text{tune} = \varnothing$ (the \emph{prefix-LM regime} on PaliGemma2), the LoRA operator has no layers to attach to: $\Delta = \varnothing$, and $\mathcal{L}_\text{task}(\Theta', \varnothing) = \mathcal{L}_\text{task}(\Theta', \mathbf{0})$ is constant in the optimisation variables. The total objective therefore degenerates to
\begin{equation}
    \mathcal{L}_\text{total}^\text{(2-op)}(\Theta')
    \;=\; \lambda_\text{MMD} \cdot \mathcal{L}_\text{MMD}(\mathcal{B}_s, \mathcal{B}_t),
    \label{eq:ltotal_2op}
\end{equation}
i.e.\ CISR automatically becomes MMD-on-$\Phi_\text{align}$\,+\,freeze-on-$\Phi_\text{freeze}$, without any user-side recipe change. Concretely, $|\mathcal{O}_\text{active}| = 3$ on causal-attention VLMs and $|\mathcal{O}_\text{active}| = 2$ on prefix-LM VLMs; experiments in \S\ref{sec:exp:main} validate both regimes.

\subsection{Full procedure and zone-matched rationale}
\label{app:method:procedure}
The three operators are matched to the type of shift that CIS measures in each zone, and each of the three matches is necessary:
\textbf{(i) MMD on $\Phi_\text{align}$} aligns two visual-appearance distributions via a rank-$r$ LoRA on attention projections: source and target features become statistically indistinguishable on these layers, while the MLP-level feature transformation is preserved (no MLP adapter attached). Since image-channel shift is concentrated here and is high-dimensional, distributional alignment is the right tool.
\textbf{(ii) LoRA on $\Phi_\text{tune}$} (when this zone is non-empty) adds a low-rank correction that is sufficient for the numerical-distribution shift but small enough to avoid overfitting to the source. Since text-channel shift acts along a low-dimensional subspace, a parameter-efficient update is sufficient.
\textbf{(iii) Freeze on $\Phi_\text{freeze}$} preserves the task-critical invariant computation that cannot be improved by adaptation. Updating these layers (e.g., via LLRD~\citep{howard2018universal} or full LoRA) only adds noise and overfitting risk; freezing them is strictly better.
The assignment is architecture-aware by construction: the 3-operator instantiation emerges on causal decoders where all three zones are non-empty, and the 2-operator instantiation (MMD\,+\,freeze) emerges on prefix-LM decoders where the text-dominated zone collapses (Eq.~\ref{eq:ltotal_2op}). Applying any single operator uniformly across all layers is strictly worse than this matched assignment; in the extreme case of global MMD (MMD-only ablation), the result is catastrophic (\S\ref{sec:exp:ablation}).

\paragraph{Full procedure.}
Algorithm~\ref{alg:molarx} consolidates the equations from \S\ref{sec:cis}--\S\ref{sec:cisr} into a single executable procedure. Lines~1--3 implement CIS (locate); line~4 sets up CISR's per-zone parameter masks; the inner loop on lines~5--9 implements CISR (repair). On a prefix-LM VLM ($\Phi_\text{tune} = \varnothing$), line~4's LoRA-attachment step is a no-op and the $\mathcal{L}_\text{task}$ term contributes no gradient, so the algorithm transparently realises the 2-operator degeneration (Eq.~\ref{eq:ltotal_2op}) without any user-side change. Computational cost: lines~1--3 (CIS diagnosis) require $2 \cdot N \cdot L$ partial forward passes with one swap per pass, totalling $\sim 0.65$\,h on a single A10 GPU at $N{=}100, L{\approx}20$ (App.~\ref{app:n_pair}); the CISR inner loop (lines~5--9) has the same per-step cost as standard LoRA fine-tuning plus one extra forward pass per layer in $\Phi_\text{align}$ for the MMD term.

\begin{algorithm}[h]
\caption{MoLaRx: Locate-then-Repair for VLM Trajectory Planning}
\label{alg:molarx}
\begin{algorithmic}[1]
\REQUIRE fine-tuned planner $f_{\Theta'}$ (Appendix~\ref{app:tvg}); labelled source $\mathcal{S}_s$; unlabelled target $\mathcal{S}_t$; thresholds $\tau_\text{img}, \tau_\text{txt}$; weight $\lambda_\text{MMD}$; paired-sample budget $N$.
\STATE \textbf{CIS, sample.} Draw $N$ scene-matched pairs from $\mathcal{P}_\text{paired}$ (Definition~\ref{def:paired}).
\STATE \textbf{CIS, score.} For each $(m, l) \in \{\text{img}, \text{txt}\} \times \{0, \ldots, L{-}1\}$, compute $\widehat{\boldsymbol{\Psi}}[m, l]$ via Eq.~\ref{eq:cis}.
\STATE \textbf{CIS, partition.} Form $\Phi_\text{align}, \Phi_\text{tune}, \Phi_\text{freeze}$ via Eqs.~\ref{eq:phi_align} to \ref{eq:phi_freeze}.
\STATE \textbf{CISR, setup.} Attach rank-$r$ LoRA on $\Phi_\text{align}$ (attention only, MMD-driven) and on $\Phi_\text{tune}$ (all projections, task-driven) per Eq.~\ref{eq:lora}; zero-mask all gradients on $\Phi_\text{freeze}$ (Eq.~\ref{eq:grad_flow}).
\FOR{each training step}
    \STATE Sample labelled batch $\mathcal{B}_s \sim \mathcal{S}_s$ and unlabelled batch $\mathcal{B}_t \sim \mathcal{S}_t$.
    \STATE Compute $\mathcal{L}_\text{task}(\Theta', \Delta;\, \mathcal{B}_s)$ (Eq.~\ref{eq:l_task}) and $\mathcal{L}_\text{MMD}(\Theta';\, \mathcal{B}_s, \mathcal{B}_t)$ (Eq.~\ref{eq:l_mmd}).
    \STATE Backpropagate $\mathcal{L}_\text{total}$ (Eq.~\ref{eq:ltotal}) with the per-zone gradient masking of Eq.~\ref{eq:grad_flow}.
\ENDFOR
\RETURN domain-adapted planner $f^* := f_{\Theta'+\Delta^*}$.
\end{algorithmic}
\end{algorithm}


\end{document}